\pdfoutput=1

\documentclass[11pt]{article}

\usepackage[final]{acl}

\usepackage{times}
\usepackage{latexsym}
\usepackage{setspace}
\usepackage[breakable]{tcolorbox}
\usepackage{listings}
\usepackage{xcolor}
\usepackage{tcolorbox}
\usepackage{listings}
\usepackage{xcolor}
\tcbuselibrary{listings}
\tcbuselibrary{breakable}
\tcbuselibrary{skins}
\newtcolorbox{prompt}[1]{
    enhanced,
    drop shadow=black!5!white,
    left=4mm,
    right=4mm,
    top=2mm,
    bottom=2mm,
    boxsep=0mm,
    rounded corners,
    title=#1,
    fontupper=\footnotesize\linespread{0.9}\fontfamily{lmr}\selectfont,
    breakable 
    }
\definecolor{lightblue}{rgb}{0.68, 0.85, 0.90}
\usepackage[T1]{fontenc}

\usepackage[utf8]{inputenc}

\usepackage{microtype}

\usepackage{inconsolata}

\usepackage{graphicx}

\usepackage{booktabs}
\usepackage{tcolorbox}
\usepackage{graphicx}
\usepackage{hyperref}
\usepackage{float}
\usepackage{natbib}
\usepackage[utf8]{inputenc}
\usepackage{newunicodechar}
\DeclareUnicodeCharacter{1F99C}{\LaTeX}

\usepackage{cite} 
\usepackage{xcolor} 
\usepackage{amsmath}
\usepackage{lipsum}
\usepackage{multirow}
\usepackage{array}  
\usepackage{ragged2e}  

\definecolor{mydarkblue}{HTML}{012E4F}
\definecolor{mylightblue}{HTML}{C2D2EF}
\definecolor{nodeblue}{HTML}{0D31D9}

\newcommand{\method}[0]{{\textsc{Storyteller}}}

\newcommand{\storyline}[0]{{\textsc{Storyline}}}
\newcommand{\nekg}[0]{\textsc{NEKG}}
\newcommand{\CBN}[0]{\textit{CBN}}
\newcommand{\CEN}[0]{\textit{CEN}}
\newcommand{\CPN}[0]{\textit{CPN}}

\title{\method: An Enhanced Plot-planning Framework \\ for Coherent and Cohesive Story Generation}

\author{
Jiaming Li$^{1,2}$\footnotemark[1] \quad
Yukun Chen$^{1,2}$\footnotemark[1] \quad
Ziqiang Liu$^{1,2}$ \quad
Minghuan Tan$^{1}$\quad
Lei Zhang$^{1,2}$ \quad
\\
\textbf{Yunshui Li}$^{1,2}$ \quad
\textbf{Run Luo}$^{1,2}$ \quad
\textbf{Longze Chen}$^{1,2}$ \quad
\textbf{Jing Luo}$^{1}$ \quad
\textbf{Ahmadreza Argha}$^{3}$
\\
\textbf{Hamid Alinejad-Rokny}$^{3}$  \quad
\textbf{Wei Zhou}$^{4}$\footnotemark[2]  \quad
\textbf{Min Yang}$^{1}$\\
\textsuperscript{1}Shenzhen Institutes of Advanced Technology, Chinese Academy of Sciences\\
\textsuperscript{2}University of Chinese Academy of Sciences\\
\textsuperscript{3}UNSW Sydney, Sydney, NSW, Australia\;
\textsuperscript{4}Chongqing University\\
\texttt{\{jm.li4, yk.chen2, min.yang\}@siat.ac.cn}
}

\begin{document}
\maketitle

\renewcommand{\thefootnote}{\fnsymbol{footnote}}
\footnotetext[1]{Equal contribution.}
\footnotetext[2]{Corresponding author.}
\renewcommand{\thefootnote}{\arabic{footnote}}

\begin{abstract}
	Stories are central to human culture, serving to share ideas, preserve traditions, and foster connections.
	Automatic story generation, a key advancement in artificial intelligence (AI), offers new possibilities for creating personalized content, exploring creative ideas, and enhancing interactive experiences. However, existing methods struggle to maintain narrative coherence and logical consistency. This disconnect compromises the overall storytelling experience, underscoring the need for substantial improvements.
	Inspired by human cognitive processes, we introduce \method{}, a novel approach that systemically improves the coherence and consistency of automatically generated stories. \method{} introduces a plot node structure based on linguistically grounded subject-verb-object (SVO) triplets, which capture essential story events and ensure a consistent logical flow. Unlike previous methods, \method{} integrates two dynamic modules—the \storyline{} and narrative entity knowledge graph (\nekg)—that continuously interact with the story generation process. This integration produces structurally sound,  cohesive and immersive narratives.
	Extensive experiments demonstrate that \method{} significantly outperforms existing approaches, achieving an 84.33\% average win rate through human preference evaluation. At the same time, it is also far ahead in other aspects including creativity, coherence, engagement, and relevance.

\end{abstract}

\section{Introduction}

\begin{quotation}
	\noindent\textit{"Stories make us more alive, more human, more courageous, more loving."}

	\hfill --- \textit{Madeleine L'Engle}
\end{quotation}

\begin{figure}[!t]
	\centering
	\includegraphics[width=\columnwidth]{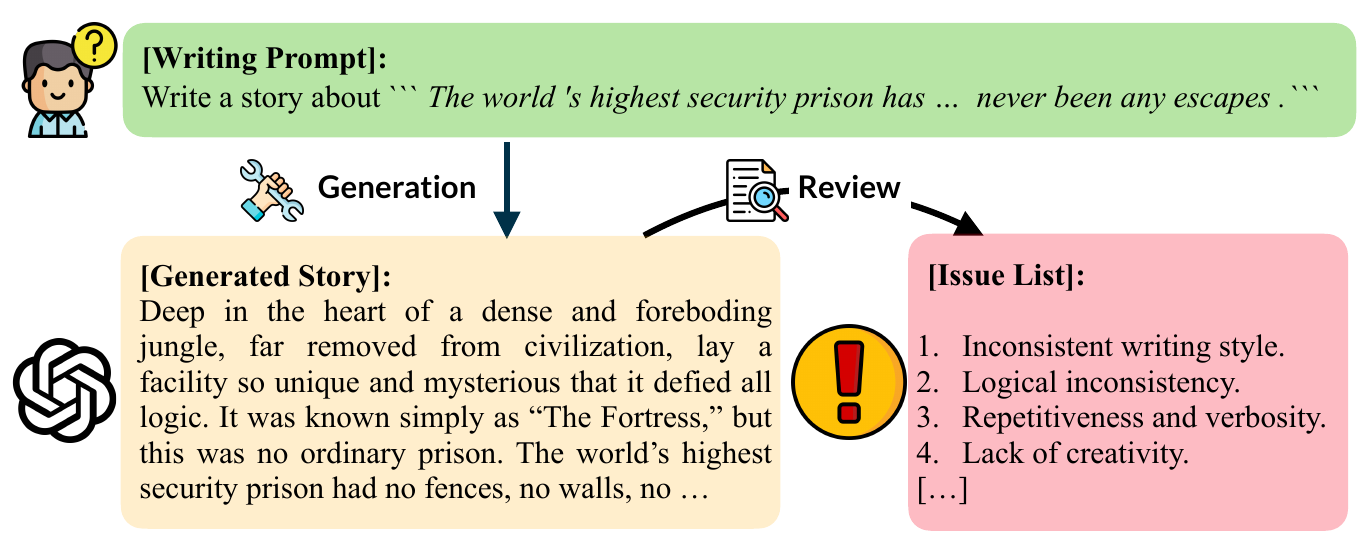}
	\caption{Story generation still faces challenges like inconsistent style, logic issues, lack of creativity and so on.}
	\label{fig:Example}
\end{figure}

Stories have always been integral to human society, serving as a means to share knowledge, entertain, and foster deeper emotional connections \citep{Cupitt1991-CUPWIA}. From ancient oral traditions to modern digital media, storytelling shapes our understanding of the world and strengthens communal bonds. Advances in artificial intelligence have positioned automatic story generation \citep{alhussain2021automatic,alabdulkarim2021automatic} as a pivotal  frontier in natural language processing (NLP), offering personalized educational content and enhancing user experiences in interactive systems. While large language models (LLMs) \citep{achiam2023gpt,touvron2023llama,zhang2022opt} excel at generating stylistically fluent text, they frequently falter in maintaining coherence and logical consistency, particularly in preserving plot structure and thematic continuity across extended narratives.

Existing hierarchical and event-based generation methods rely on high-level outlines to guide narrative development \citep{fan2018hierarchical,yao2019plan,rashkin2020plotmachines}. However, these methods often generate chapters or events independently, resulting in plot inconsistencies and fragmented narrative cohesion. For example, story segments generated without contextual awareness may introduce abrupt shifts in character motivations or jarring plot deviations that undermine narrative immersion \citep{yao2019plan}. Such limitations fundamentally restrict the capacity of these systems to produce engaging, structurally sound stories.

To address these challenges, we draw inspiration from the cognitive process of human writers \citep{lamott1995bird,cook2011plotto}. Consider a writer crafting a complex story: they continuously retrieve past events, evaluate how new developments align with established narrative elements, and generate content that integrates seamlessly into the evolving plot \citep{clark2008writing,brande2013becoming}. This cognitive loop—Retrieval, Evaluation, and Generation—ensures natural  story evolution, where each  event logically builds on prior context.

Building on this insight, our proposed framework, \method{}, introduces an adaptive plot-planning mechanism for story generation. Unlike prior methods, \method{} enables continuous interaction among plot points during generation, ensuring cohesive narrative structure. Central to our approach is a dynamic plot node mechanism that tracks key events and character interactions through subject-verb-object (SVO) triplets, augmented by a storyline and narrative entity knowledge graph. This design preserves logical flow and thematic consistency while offering granular control over plot progression, enhancing both coherence and depth.

Highlights of our contributions are as follows:

\begin{itemize}
	\item A novel plot node structure using SVO triplets to capture story events, ensuring logical progression and thematic continuity, key to coherent long-form narratives.
	\item Two dynamically integrated modules: the \storyline{} and narrative entity knowledge graph (\nekg), which interact bidirectionally with the generation process to systematically refine plot structure and event coherence across narrative stages.
	\item Empirical validation demonstrating that \method{}  significantly outperforms existing models, generating stories with superior coherence, creativity, engagement, and overall quality.
\end{itemize}

\section{Related Work}

\paragraph{Large Language Models.}

The emergence of LLMs has revolutionized NLP \citep{vaswani2017attention, brown2020language, zhang2023marathon}. Early architectures like BERT \citep{devlin-etal-2019-bert} and the Transformer \citep{vaswani2017attention} laid the groundwork for  advanced successors such as GPT-4 \citep{achiam2023gpt}, Llama-3 \citep{llama3modelcard}, and Qwen \citep{qwen}, which are distinguished  by their scalability and versatility. These models serve as foundational platforms for diverse applications, excelling in tasks like machine translation, text summarization, and conversational AI. Their capacity to generate coherent and contextually relevant text has driven innovations across domains including healthcare, education, finance, and story generation

\paragraph{Automatic Story Generation.}

Researchers have long sought to develop systems capable of automatically generating stories \citep{wang2023open}. Early studies, such as Tale-spin \citep{meehan1977tale} and Minstrel \citep{turner1993minstrel}, generated stories through predefined rules and structured frameworks. However, these approaches were limited by closed-world settings, lacking flexibility and scalability. \citep{tearse2010minstrel} improves upon Minstrel and explores how modern technologies can enhance the flexibility of story generation. \citep{li2013story} extracts events from human-written plots and constructs a causal relationship graph between the events.

\begin{figure*}[!t]
	\centering
	\includegraphics[width=\textwidth]{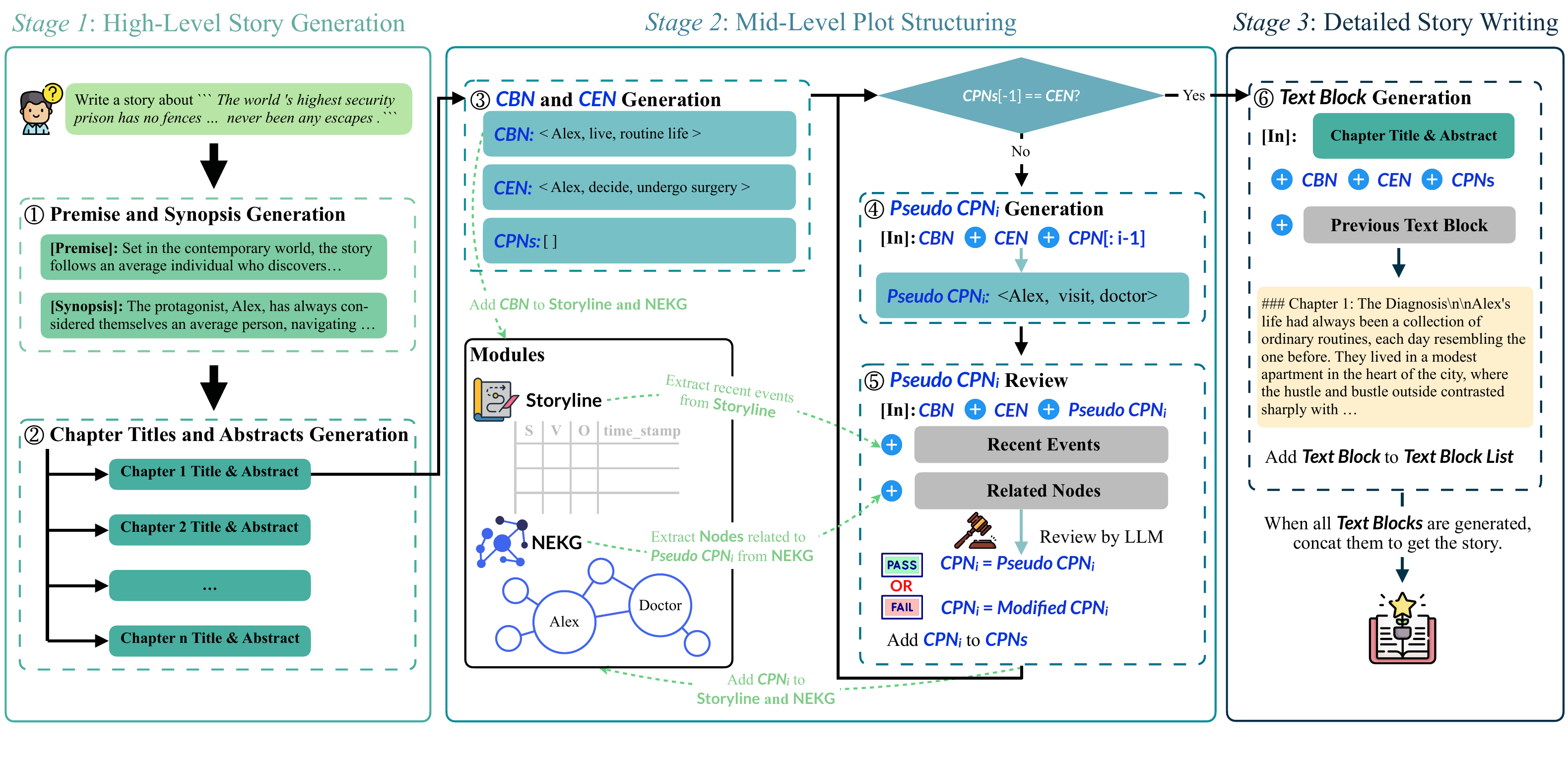}
	\caption{Overview of \method. The method is divided into three stages. This figure illustrates the complete process by which \method{} transforms user input into the final story. Additionally, it highlights the information interaction between various stages and both the \storyline{} and \nekg{} module.}
	\label{fig:method}
\end{figure*}

In recent years, neural networks have been widely used for story generation \citep{fan2018hierarchical}. Hierarchical story generation has become a mainstream approach, where methods like \citep{yao2019plan, yang2022re3} first generate high-level outlines and then expand them into chapters or paragraphs. Dramatron \citep{mirowski2023co} leverages LLMs combined with this approach to generate scripts and screenplays. DOC \citep{yang2022doc} generates more coherent long stories through more detailed planning and stronger dynamic control. \citet{guan2020knowledge} introduces external commonsense knowledge bases and integrates multi-task learning to enhance the model's ability to capture causal relationships and temporal dependencies. CAST~\citep{peng-etal-2022-inferring} also use commonsense knowledge to guide story generation. StoRM \citep{peng2021guiding} models readers' understanding of the story world in the form of knowledge graphs. However, narrative coherence and structural consistency remain challenges. We propose a method to enhance the narrative coherence and logical consistency of stories through a plot node mechanism based on Subject-Verb-Object (SVO) and dynamic interaction modules ( \storyline{} and \nekg{}).

\paragraph{Prompting.}

Prompting leverages textual instructions to guide LLM outputs \citep{liu2023pre}. A cornerstone of LLM versatility is in-context learning, which enables task execution based solely on prompt content without requiring parameter updates \citep{dong2022survey, wei2022emergent}. This capability grants LLMs exceptional adaptability, allowing them to generalize across diverse tasks with minimal fine-tuning \citep{levine2022the, zhao2021calibrate}.

\section{Methodology}

This section describes the architecture of \method{} and the interactions amongits components. First, we provide an overview of three core modules ($\S$\ref{subsec:Modules}). We then present a top-down workflow analyzing the sequential stages of narrative generation: \textit{Stage 1} ($\S$\ref{subsec:Stage 1}), \textit{Stage 2} ($\S$\ref{subsec:Stage 2}), and \textit{Stage 3} ($\S$\ref{subsec:Stage 3}).

\subsection{Modules}
\label{subsec:Modules}

\paragraph{\textsc{Nodes}.}

In \method, the core structural element is the \textsc{Node}. Each \textsc{Node} represents either a SVO or a subject-verb (SV). To standardize representation, SV pairs are converted into subject-verb-subject (SVS) triplets (Equation \ref{eq:Node SV}).

\begin{align}
	\label{eq:Node SVO}
	 & \textsc{Node}({\mathrm{SVO}})= <\mathrm{Subject}, \mathrm{Verb}, \mathrm{Object}> \\
	\label{eq:Node SV}
	 & \textsc{Node}({\mathrm{SV}})= <\mathrm{Subject}, \mathrm{Verb}, \mathrm{Subject}>
\end{align}

\textsc{Nodes} serves as the interface between \storyline{} and \nekg{} enabling the integration of high-level information into the story generation process. Furthermore, \textsc{Nodes} are classified into three types based on their position within a chapter: \CBN, \CPN{} and \CEN{} (Table \ref{tab:Nodes}).

\begin{table}[!t]
	\centering
	\scalebox{1.0}{
		\begin{tabular}{l|p{5cm}}
			\toprule
			\textbf{\textsc{Node}} & \textbf{Description}                                    \\
			\midrule
			\CBN                   & The start node of a chapter                             \\
			\midrule
			\CPN                   & Plot node within the chapter, driving the story forward \\
			\midrule
			\CEN                   & The end node of a chapter                               \\
			\bottomrule
		\end{tabular}
	}
	\caption{\label{tab:Nodes}
		Three \textsc{Node} types: \CBN{}(Chapter Begin Node), \CPN{}(Chapter Plot Node) and \CEN{}(Chapter End Node). These \textsc{Nodes} represent stages in the narrative structure of a chapter.
	}
\end{table}

\begin{figure*}[!t]
	\centering
	\includegraphics[width=\textwidth]{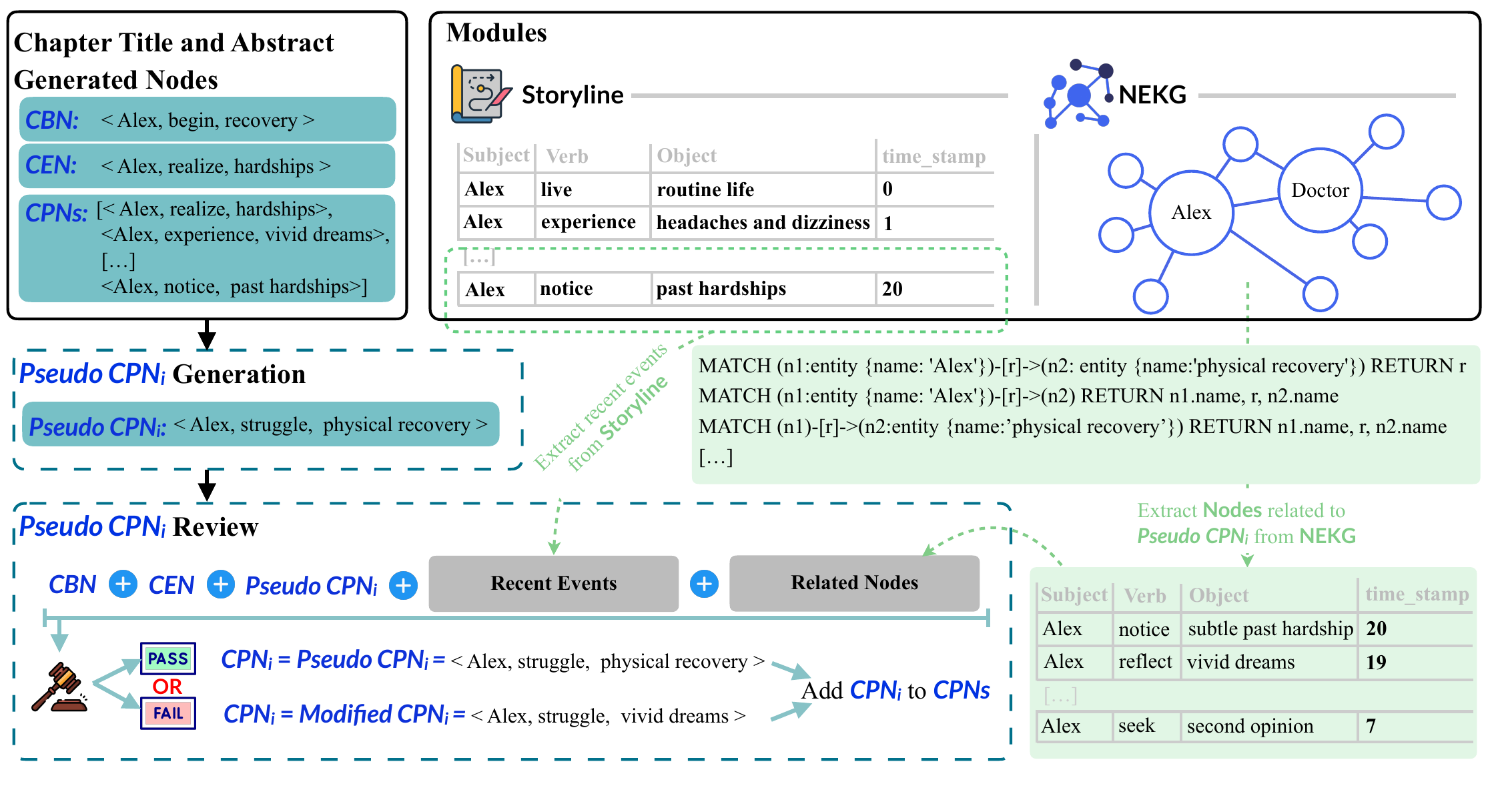}
	\caption{\textit{Pseudo CPN} Generation and Review. Specific example is used to provide a detailed description of the process. Additionally, it is clearly explained how the \storyline{} and \nekg{} provide relevant information to aid in the generation of \CPN.}
	\label{fig:CPN}
\end{figure*}

\paragraph{\storyline.}

\storyline{} stores the \textsc{Nodes} generated during the story generation process, tracking the development of the narrative plot.

Each \textsc{Node} is assigned a \texttt{time\_stamp} indicating temporal position, enabling the establishment of a chronological sequence across \textsc{Node}. This ensures accurate event ordering and captures the progression of the narrative. By maintaining these temporal relationships, \storyline{} provides a structured record of the evolving plot, which is essential for preserving story coherence and understanding the interconnections between events and actions over time.

\paragraph{\nekg.}

The narrative entity knowledge graph (\nekg) represents entities and their interrelationships in a structured, graph-based format. It comprises nodes and edges, where nodes represent entities or concepts (e.g., characters, locations, objects), and edges denote the relationships between them. \nekg{} effectively captures character interactions, object dynamics, and the logical progression of events within a story.

\nekg{} serves as a tool for both systems and readers to intuitively understand the key drivers of narrative progression and uncover implicit  relational chains. We implement \nekg{} using Neo4j\footnote{\url{https://neo4j.com/}}, with each generated story having a corresponding \nekg instance.

In addition, \nekg{} is highly extensible. Beyond describing the current storyline, it supports future plot generation and reasoning. By leveraging rich associative data and logical relationships, the graph enables narrative expansion based on existing information, facilitating the creation of new plotlines or character arcs. This capability not only aids story analysis but also inspires and guides creative development.

\subsection{\textit{Stage 1}: High-Level Story Generation}
\label{subsec:Stage 1}

In this stage, \method{} generates the story at a high-level. First, it consolidates user-provided information to produce a premise and synopsis. Next, to ensure coherence and smooth flow, it generates titles and abstracts for all chapters.

\paragraph{Premise and Synopsis Generation.}

This phase transforms user input into two key components: a concise premise and a detailed synopsis. The input is analyzed to extract core elements, which are used to generate these components. The premise  describes the story's era, setting, and societal context, while the synopsis outlines the main plot, character dynamics, and key narrative twists.

\paragraph{Chapter Titles and Abstracts Generation.}

This phase structures the long-form story into chapters based on its premise and synopsis, generating titles and detailed abstracts for each chapter. This process establishes a comprehensive story framework, ensuring that chapter division and content creation align with the story's background and core narrative. Each chapter is assigned a relevant title and a summary detailing key events, character interaction, and plot developments to advance the storyline.

\subsection{\textit{Stage 2}: Mid-Level Plot Structuring}
\label{subsec:Stage 2}

In this stage, \method{} generates \textsc{Nodes} for each chapter. Both Chapter Begin Node (\CBN) and Chapter End Node (\CEN) are created first, followed by the sequential generation of Chapter Plot Nodes (\CPN) in two steps.

\paragraph{\CBN{} and \CEN{} Generation.}

This task involves generating \CBN{} and \CEN{} for each chapter. The process reviews the chapter's title and abstract, ensuring that the generated \CBN{} and \CEN{} maintain logical coherence and smooth transition adjacent chapters. These  \textsc{Nodes} represent the chapter's beginning and conclusion, so they incorporate information from the current, preceding, and following chapters.

\paragraph{\textit{Pseudo CPN} Generation and Review.}

The previous module's design enables structured storage of generated story information, providing a foundation for \textit{CPN} generation to ensure plot quality. However, due to context length and cost constraints of LLMs, only a subset of data from \storyline{} and \nekg{} can be used, raising the challenge: {\it How to extract relevant information from the extensive data?}

To address the, \textit{Pseudo CPN} is generated based on existing \CBN, \CEN{} and \CPN. Using the \textit{Pseudo CPN} ($\mathrm{S}^\prime\mathrm{V}^\prime\mathrm{O}^\prime$), relevant nodes matching $\mathrm{S}^\prime$ and $\mathrm{O} ^\prime$ are retrieved from \nekg, representing the most recent events between $\mathrm{S}^\prime$ and $\mathrm{O} ^\prime$. Nodes where the subject or object are $\mathrm{S}^\prime$ or $\mathrm{O}^\prime$ are also retrieved, representing recent events related to $\mathrm{S}^\prime$ and $\mathrm{O}^\prime$ (Figure \ref{fig:CPN}).

These nodes are reviewed to assess whether the \textit{Pseudo CPN} maintains coherence and fluency. If it passes, the \textit{Pseudo CPN} is accepted as the \CPN. If it fails, LLM generates a \textit{modified CPN} based on contextual information and the \textit{Pseudo CPN}, along with an explanation for the modification. The \CPN{} is added to \textit{CPNs}, and this process repeats until the generated \CPN{} aligns with \CEN, signaling the completion of the chapter's plot node generation.Details of CPN's review can be found in Section~\ref{appendix:CPN' Review Display}.

\subsection{\textit{Stage 3}: Detailed Story Writing}
\label{subsec:Stage 3}

\paragraph{\textit{Text Block} Generation.}

In this stage, the model adheres strictly to the established plot nodes (\CBN, \textit{CPNs}, \CEN) to ensure narrative coherence and consistency. The precise plot structure maintains the logical flow of events. Additionally, to ensure consistent linguistic style across the narrative, the previously generated text block from the preceding chapter is also input. This approach ensures alignment with the overall narrative style and structural requirements of the story.

\begin{table}[t]
	\centering
	\begin{tabular}{ll}
		\midrule
		\# Train Stories          & 272,600 \\
		\# Test Stories           & 15,138  \\
		\# Validation Stories     & 15,620  \\
		\midrule
		\# Prompt Words           & 7.7M    \\
		\# Story Words            & 200M    \\
		\midrule
		Average Length of Prompts & 28.4    \\
		Average Length of Stories & 734.5   \\
		\midrule
	\end{tabular}

	\caption{\label{tab:Statiscs of WritingPrompts dataset}
		Statiscs of \textsc{WritingPrompts} dataset.
	}
\end{table}

\section{Experiments}

\begin{table*}[h]
	\centering
	\scalebox{0.91}{
		\begin{tabular}{lcccccc}
			\toprule
			\multirow{2}{*}{\textbf{Model / Method}} & \textbf{Avg}        & \multicolumn{4}{c}{\textbf{Perspectives}} & \multirow{2}{*}{\textbf{Overall}}                                                 \\
			\cmidrule(lr){3-6}
			                                         & \textbf{Word Count} & Creativity                                & Coherence                         & Engagement    & Relevance                     \\
			\midrule
			\vspace{-0.3em}\textit{Model}            &                     &                                           &                                   &               &               &               \\
			\midrule
			GPT-4o                                   & 1,076               & 50.0                                      & 50.0                              & 50.0          & 50.0          & 50.0          \\
			Qwen2-72B-Instruct                       & 5,173               & 25.0                                      & 25.0                              & 25.5          & 31.3          & 18.3          \\
			Llama3.1-70B-Instruct                    & 1,578               & 46.2                                      & 50.8                              & 42.4          & 49.2          & 31.4          \\
			LongWriter-glm4-9b                       & 5,191               & 60.2                                      & \underline{68.3}                              & \underline{64.0}          & \underline{67.3}          & \underline{62.7}          \\
			LongWriter-llama3.1-8b                   & 5,659               & 47.8                                      & 53.5                              & 51.8          & 54.7          & 43.1          \\
			\midrule
			\vspace{-0.3em}\textit{Method}           &                     &                                           &                                   &               &               &               \\
			\midrule
			DOC v2                                   & \underline{7,323}               & \underline{60.4}                                      & 49.7                              & 52.2          & 51.6          & 50.9          \\
			\method{} (ours)                         & \textbf{7,594}      &  \textbf{74.7}                           &                    \textbf{72.8} & \textbf{66.5} & \textbf{77.6}& \textbf{89.4} \\
            \;\;- w/o \nekg                        & 7,459      &                            72.2  &                     67.8 & 64.2 & 67.3 & 87.7 \\
            \;\;- w/o \textit{Pseudo CPN} Review                    & 5,204      &                         68.8    &               67.7       & \textbf{66.5} & 71.0 & 79.1 \\
			\bottomrule
		\end{tabular}
	}
	\caption{\label{tab:Comprehensive Evaluation Results}
		Overall results of different models or methods. \method{} excels across all metrics, achieving a high average word count (7,594) and the highest Overall score of 89.4. This performance significantly outperforms other models. Additionally, \method{}  leads in five distinct evaluation perspectives. The best-performing model for each metric is highlighted \textbf{in-bold}, and the second-best model is \underline{underlined}.
	}
\end{table*}

\subsection{Experiments Setup}

\paragraph{Dataset.}
To evaluate the performance of \method, we conduct experiments using the \textsc{WritingPrompts} dataset \citep{fan2018hierarchical}. This dataset contains human-written stories and their corresponding prompts, sourced from Reddit\footnote{\url{www.reddit.com/r/WritingPrompts/}}. The prompts span a wide range of themes. To balance evaluation cost and ensure comprehensive assessment, we randomly select 200 prompts tagged with \texttt{[ WP ]} from the dataset. A detailed description of \textsc{WritingPrompts} is provided in Appendix \ref{appendix:WritingPrompts Dataset}. Moreover \citet{fan2018hierarchical} analyze the statistics of the \textsc{WritingPrompts} dataset, and here we directly reference their work as shown in Table \ref{tab:Statiscs of WritingPrompts dataset}.

\paragraph{Baselines.}

Prior to this work, most models or methods focused on generating shorter stories compared to \method. Therefore, we select the following baselines. The prompt template is provided in Appendix \ref{appendix:Chat Template}.

\begin{itemize}
	\item \textbf{GPT-4o}\footnote{GPT-4o version(all experiments): 2024-08-06} \citep{HelloGPT4o}: OpenAI's most advanced GPT model, designed for complex, multi-step tasks.
	\item \textbf{Qwen2-72B-Instruct} \citep{qwen} and \textbf{Meta-Llama-3.1-70B-Instruct} \citep{llama3modelcard}: Two open-source instruction-tuned models.
	\item \textbf{LongWriter Models} \citep{bai2024longwriter}: Based on Meta-Llama-3.1-8B \citep{dubey2024llama} and GLM4-9B \citep{glm2024chatglm}, capable of generating over 10,000 words in a single output.
	\item \textbf{DOC v2}\footnote{\url{https://github.com/facebookresearch/doc-storygen-v2}}: An improved version of DOC, featuring a detailed outliner for structured, multi-level outlines and a controller to ensure adherence to the outline.
\end{itemize}

\paragraph{Configuration.} All open-source models are inferred on 8 * NVIDIA A100 GPUs using \texttt{vllm}\footnote{An efficient LLM inference engine} \citep{kwon2023efficient}, with \texttt{temperature} set to 0 and \texttt{max\_tokens} set to 16,384 in  \texttt{SamplingParams}, employing greedy decoding for inference.

For DOC v2 and \method{}, GPT-4o is utilized as the engine. Additionally, GPT-4o serves as the judge model for all evaluations.

\paragraph{Metrics.}

To comprehensively assess the performance of stories generated by \method, evaluations are conducted from multiple perspectives. First,  the length of generated texts is measured using the \textbf{Average Word Count} metric. Next, the overall quality is evaluated through an {\it Overall} score. Additionally, detailed assessments are performed across five specific dimensions:

\begin{enumerate}
\item \textbf{Creativity.} Originality of the plot and characters.
\item \textbf{Coherence.} Clarity of the narrative structure and flow.
\item \textbf{Engagement.} The extent to which the story captivates readers and sustains their emotional involvement and curiosity.
\item \textbf{Relevance.} Alignment with the theme, prompt, or background.
\item \textbf{Overall.} A holistic evaluation considering alignment with the prompt, character and plot development, reader engagement, originality, and areas for improvement.
\end{enumerate}

Following the \textit{arena-hard-auto} \citep{li2024crowdsourced} evaluation method, stories generated by GPT-4o served as a baseline, and comparisons are made with other generated stories to obtain quantified scores.

We also conduct a preference evaluation using GPT-4o as the annotator to compare \method{} with other baseline storylines.

Additionally, we assess the diversity of generated stories using two key metrics: Diverse Verbs and DistinctL-n. Diversity enhances the depth and breadth of narratives, making plots, characters, and settings more vibrant while avoiding monotony. 

\begin{itemize}
\item \textbf{Diverse Verbs} \citep{fan-etal-2019-strategies}: Analyzes verbs within the text to reflect action diversity. A higher variety of verbs indicates richer content. 
\item \textbf{DistinctL-n}: an adaptation of Distinct-n \citep{li-etal-2016-diversity} that accounts for text length. As text length increases,  repeated n-grams may also increase, potentially lowering the Distinct score. DistinctL-n, defined in Equation \ref{eq:DistinctL-n}, incorporates text length to provide a fairer assessment of diversity in longer texts. 
\end{itemize}
\begin{align}
	\label{eq:DistinctL-n}
	\mathrm{DistinctL-n} & =  \frac{\text{unique\,n-grams}}{\text{total\,n-grams}} \nonumber \\
	                     & \quad \times \left( 1 + \log(\text{word\_count}) \right)
\end{align}

\subsection{Comprehensive Evaluation}

Directly asking an LLM to rate generated stories across multiple dimensions can yield unstable results. This instability arises from the high variability of the stories and their often lengthy nature, which complicates the provision of effective few-shot guidance in the prompt. Inspired by \citet{li2024crowdsourced}, we select GPT-4o as the reference model and evaluate stories generated by other models against it with 200 randomly picked prompts across multiple dimensions. The score is defined as $\frac{1}{N}\sum_{\{i,j\}}(f(\pi_j<\pi_i)-O(\pi_j<\pi_i))^2$, more details can be found in Appendix~\ref{appendix:appDetails Of The Formula Derivation}. To ensure the model accurately understands the comparison criteria for each dimension, we develop a detailed checklist for each evaluation dimension, enabling the LLM to conduct comparative assessments effectively.

As shown in Table \ref{tab:Comprehensive Evaluation Results}, \method{} performs exceptionally well across all evaluated dimensions. It achieves an average word count of 7,594, and demonstrates significant advantages in Creativity, Coherence, Engagement, Relevance, and Overall Performance. With an Overall score of 89.4, \method{} outperforms all other models, none of which exceed 65 points. This underscores  \method's superior ability to generate high-quality, well-rounded narratives.

\begin{figure}[!t]
	\centering
	\includegraphics[width=0.9\columnwidth]{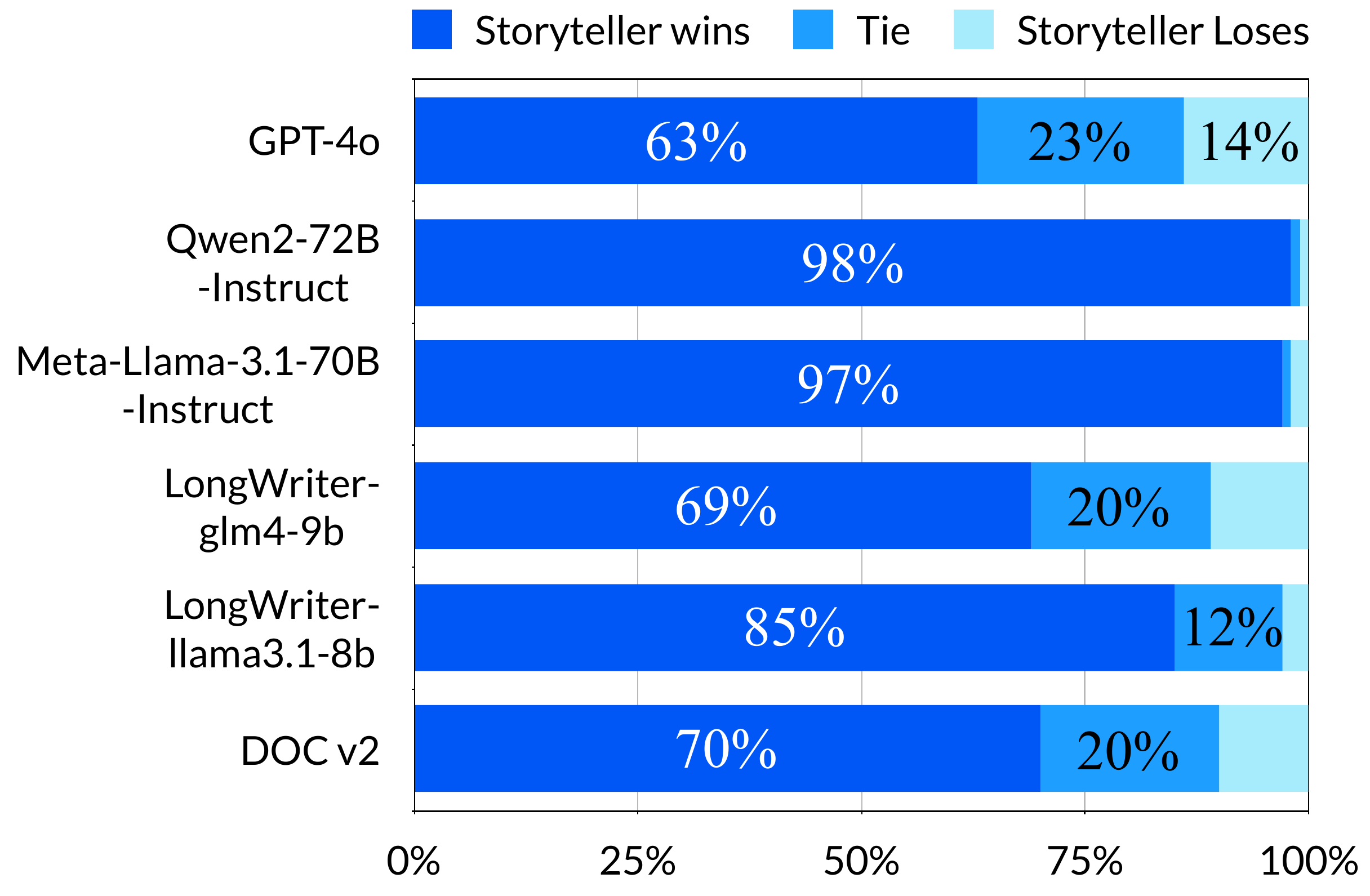}
	\caption{Results of preference evaluation using GPT-4o as the annotator. \method{} consistently outperforms all models, with over 95\% win rates against Qwen2-72B-Instruct and Meta-Llama-3.1-70B-Instruct, and less than 70\% win rates with around 20\% ties against GPT-4o and LongWriter-llama3.1-8b.
    }
	\label{fig:Preference}
\end{figure}


For Creativity, \method{} scores reaches 74.7, significantly surpassing DOC v2 (60.4), reflecting the ability to craft imaginative and innovative narratives. In Coherence, the score of \method{} is 72.8, slightly outperforming LongWriter-glm4-9b (68.3), showcasing stronger logical consistency and well-structured storytelling. For Engagement, \method{} achieves 66.5, exceeding LongWriter-glm4-9b's 64.0, demonstrating a stronger capacity to captivate audiences. In Relevance, \method{} again leads with 77.6, highlighting the capability to generate content closely aligned with prompts.

It is also noteworthy that Qwen2-72B-Instruct generates stories with a significantly higher average word count than Llama3.1-70B-Instruct; however, its subsequent metric scores are relatively lower. This discrepancy is attributed to the presence of repeated paragraphs in Qwen2-72B-Instruct's outputs, which dimishes the overall quality.

\begin{figure}[!t]
	\centering
	\includegraphics[width=\columnwidth]{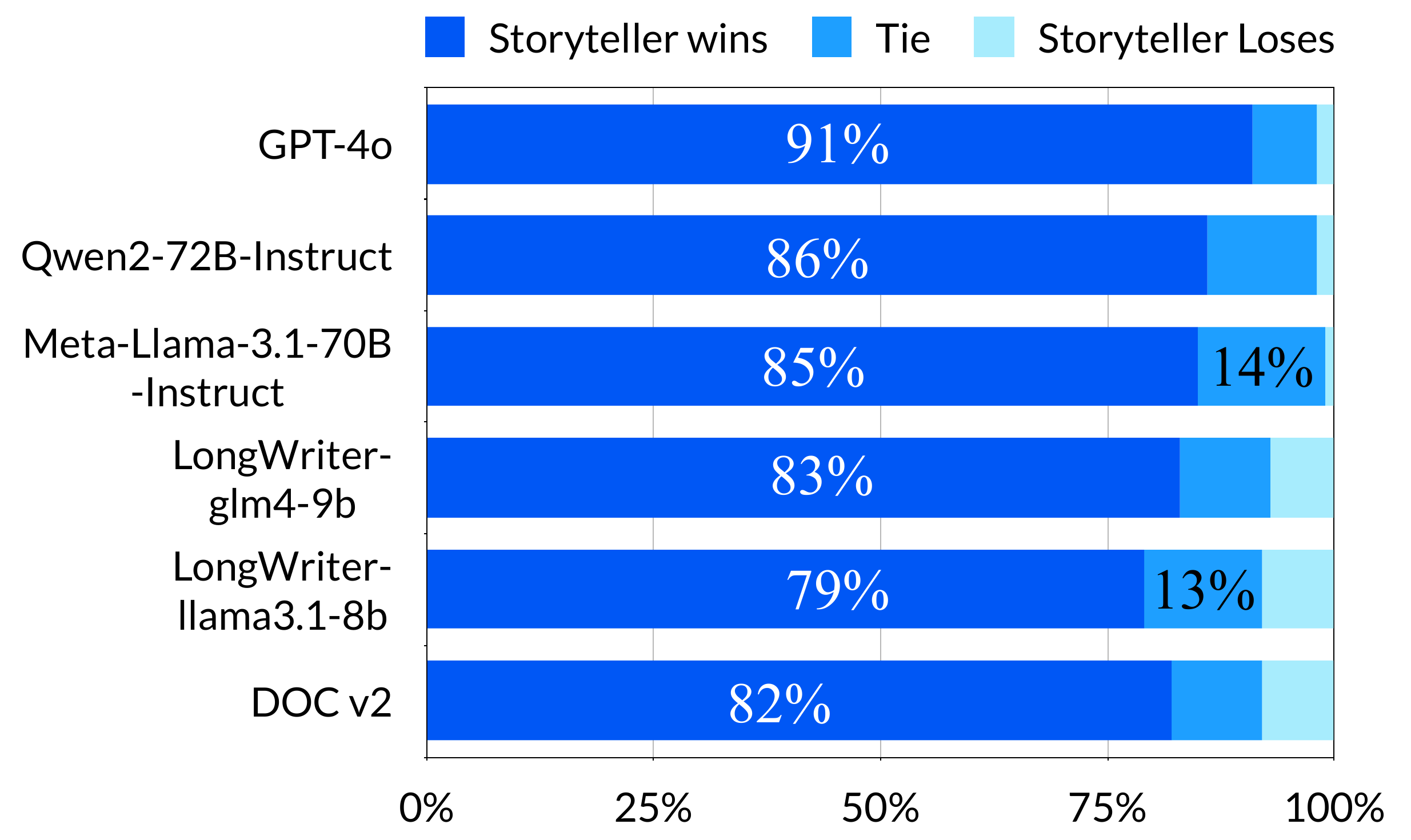}
	\caption{Results of human preference evaluation. \method{} achieves outstanding preference wins, with 79\% against LongWriter-llama3.1-8b, over 80\% against other models, including 91\% against GPT-4o and 83\% against LongWriter-glm4-9b.}
	\label{fig:Human_eval}
\end{figure}

\subsection{Preference Evaluation}

Following \citet{zhou2024lima}, we compare \method{} with other baselines by generating stories from the same set of writing prompts for all systems. To ensure a comprehensive evaluation, we employ both GPT-4o-based and human-based methods to assess the quality of the generated stories. To eliminate positional bias, the stories' positions are swapped after the first comparison, and a second round of evaluation is conducted. The final results are aggregated from both rounds.

\textbf{GPT-4o-Based Evaluation.} GPT-4o is utilized as the annotato to evaluate the output of \method{} against each baseline, determining the preferred story for every comparison.



As shown in Figure \ref{fig:Preference}, \method{} consistently outperforms all baselines in story generation. Against GPT-4o and LongWriter-glm4-9b, \method{} achieves win rates of 63\% and 69\%, respectively, with tie rates around 20\%, suggesting that while these models can occasionally match \method{}, they generally fall short in overall story quality. Compared to DOC v2, \method{} achieves a 70\% win rate with 20\% ties, reaffirming its robust performance. Notably, \method{} far surpasses the remaining models, achieving over 95\% win rates against the two Instruct models and an 85\% win rate against LongWriter-llama3.1-8b.

\begin{table}[!t]
	\centering
	\scalebox{1.0}{
		\begin{tabular}{lc}
			\toprule
			\textbf{Models/Methods} & \textbf{Cohen's Kappa}                                    \\
			\midrule
			GPT-4o        &0.8503                     \\
            Qwen2-72B-Instruct &0.8131                           \\
            Llama3.1-70B-Instruct&0.7331 \\
            LongWriter-glm4-9b&0.6603\\
            LongWriter-llama3.1-8b&0.6044 \\
            DOC v2 &0.5877\\
			\bottomrule
		\end{tabular}
	}
	\caption{\label{tab:Cohen's Kappa}
		Cohen's kappa values for inter-annotator agreement across different models and methods compared to \method.
	}
\end{table}

\textbf{Human Evaluation.} To address the biases of relying solely on GPT-4o for evaluation, we incorporate human preference assessments. Native English speakers compare pairs of generated stories using a template detailed in Appendix \ref{appendix:Human Evaluation System}. To ensure reliability, each pair is evaluated by at least two independent annotators.

As shown in Figure \ref{fig:Human_eval}, \method{} achieves strong and even better performance compared to GPT-4o-based preference evaluation. 
\method{} demonstrates outstanding preference wins across all models, achieving a 79\% win rate against LongWriter-llama3.1-8b and over 80\% against all other models. Notably, it achieves an impressive 91\% win rate against GPT-4o, the highest among all comparisons, along with 83\% against LongWriter-glm4-9b and 82\% against DOC v2.

The paper also includes an analysis of inter-annotator agreement using Cohen's kappa (Equation~\ref{kappa}) to evaluate the consistency among annotators. The results demonstrate a high level of agreement, particularly for the GPT-4o and Qwen2-72B-Instruct models, where Cohen's kappa values exceed 0.8, indicating "almost perfect agreement" according to conventional guidelines. For other models, the kappa values range from 0.58 to 0.73, reflecting "moderate to substantial agreement." These findings highlight the reliability of our human evaluation process.

\begin{align}
\label{kappa}
\kappa = \frac{P_o - P_e}{1 - P_e}
\end{align}

The slight differences between GPT-4o-based and human preference evaluations do not diminish the fact that \method{} achieves outstanding performance under both methods. These results consistently demonstrate the state-of-the-art capabilities of our model, and the two evaluation approaches together provide a comprehensive validation of its effectiveness.

\begin{figure}[!t]
	\centering
	\includegraphics[width=\columnwidth]{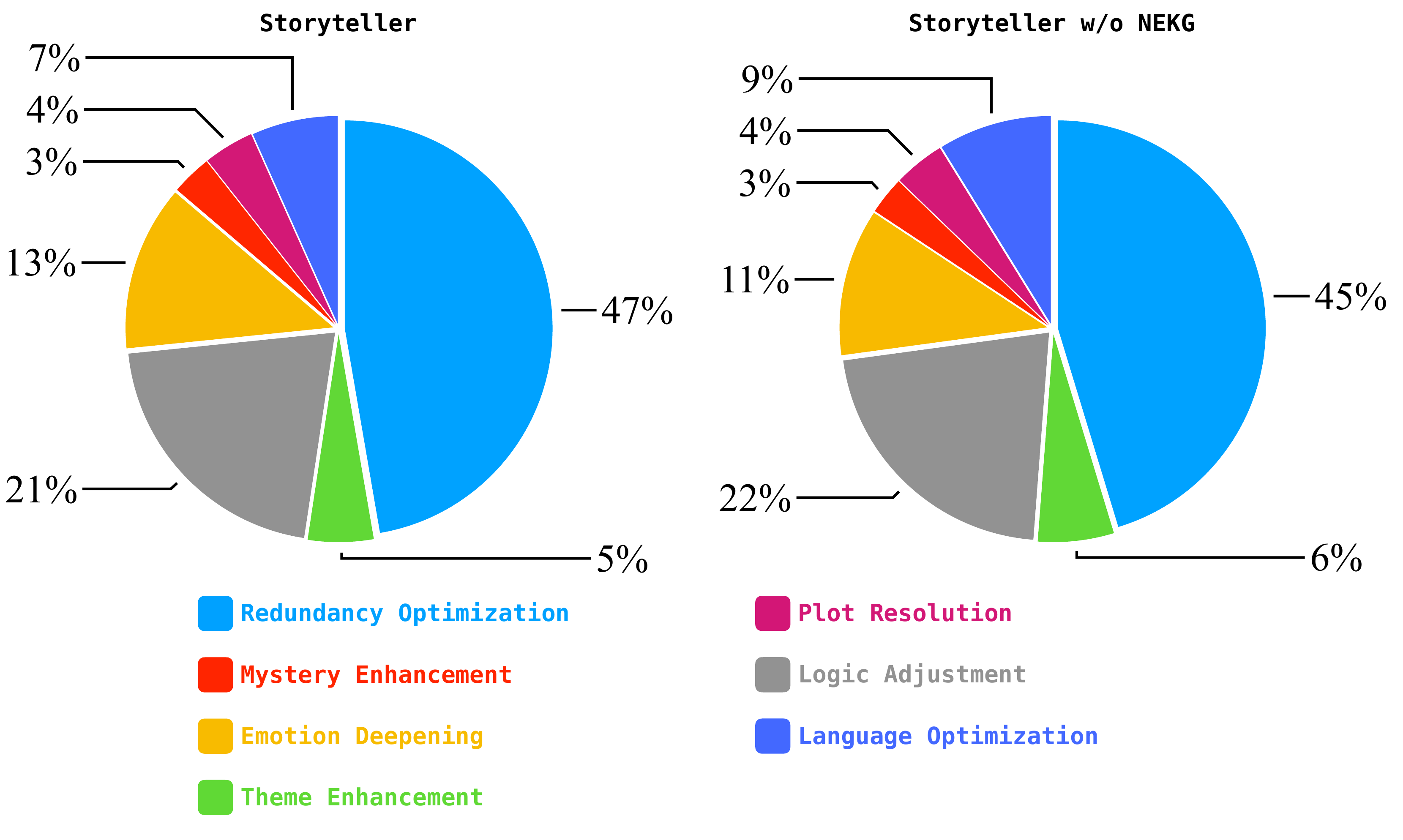}
	\caption{Proportion of CPN\_Review Types. The figure illustrates shifts in review proportions after applying \nekg, with reduced Language Optimization and Theme Enhancement, and increased Redundancy Optimization and Emotion Deepening.}
	\label{fig:CPN_Check}
\end{figure}

\begin{table*}[h]
	\centering
	\scalebox{1.0}{
		\begin{tabular}{lccccc}
			\toprule
			\multirow{2}{*}{\textbf{Model / Method}} & \multicolumn{4}{c}{\textbf{DistinctL-n}} & \textbf{Diverse}                                                             \\
			\cmidrule(lr){2-5}
			                                         & DistinctL-2                              & DistinctL-3       & DistinctL-4       & DistinctL-5       & \textbf{Verbs}   \\
			\midrule
			\vspace{-0.3em}\textit{Model}            &                                          &                   &                   &                   &                  \\
			\midrule
			GPT-4o                                   & 2.640                                    & 2.847             & 2.870             & 2.874             & \underline{0.96} \\
			Qwen2-72B-Instruct&1.572&1.888&2.004&2.064&0.83\\
			Meta-Llama3-3.1-70B-Instruct&2.297&2.692&2.820&2.879&0.91\\
			LongWriter-glm4-9b                       & 2.302                                    & 2.963             & 3.196             & 3.303             & 0.92             \\
			LongWriter-llama3.1-8b                   & 1.905                                    & 2.417             & 2.596             & 2.679             & 0.92             \\
			\midrule
			\vspace{-0.3em}\textit{Method}           &                                          &                   &                   &                   &                  \\
			\midrule
			DOC v2                                   & \textbf{3.469}                           & \underline{3.669}    & \underline{3.696} & \underline{3.703} & \textbf{0.97}    \\
			\method{} (ours)                         & \underline{3.170}                        & \textbf{3.764} & \textbf{3.856}    & \textbf{3.870}    & 0.95             \\
			\bottomrule
		\end{tabular}
	}
	\caption{\label{tab:Diversity Evaluation Results}
		Diversity results of different models or methods. \method{} excels across all metrics, achieving top rankings in diversity, particularly with the highest DistinctL-4 and DistinctL-5 scores. The best-performing model in each metric is highlighted \textbf{in-bold}, and the second best is \underline{underlined}.
	}
\end{table*}

\subsection{Ablation Study}

From Table \ref{tab:Comprehensive Evaluation Results}, after removing \nekg{} module, the overall score drops by 1.7, primarily due to a significant decrease in relevance (- 10.3) and coherence (- 5.0). The precision and logical consistency of the generated content deteriorate. On the other hand, removing the Pseudo CPN Review module results in a more substantial decline in the overall score (- 10.3), along with a reduction of 2,390 words in the average length. The content's richness and creativity are severely affected.

We select 50 stories for an analysis of the proportion of CPN\_Review types. As shown in Figure \ref{fig:CPN_Check}, Redundancy Optimization and Logic Adjustment remain the main challenges in story generation. After applying \nekg module, the proportions of Language Optimization and Theme Enhancement decrease, indicating that the story's core ideas and language can be effectively improved. The proportions of Redundancy Optimization and Emotion Deepening increase, suggesting that some information requiring consistency adjustments and enriching character development is introduced during the generation process.

\subsection{Diversity Evaluation in Story Generation}

Table \ref{tab:Diversity Evaluation Results} presents diversity metrics for all models and methods, including DistinctL-2 through DistinctL-5 and Diverse Verbs.

Overall, \method{} ranks among the top two across all DistinctL-n metrics, demonstrating strong diversity in both short and long sequences. It achieved the highest scores on DistinctL-3 (3.764), DistinctL-4 (3.856), and DistinctL-5 (3.870), reflecting its high creativity and richness in story generation. Furthermore, \method{} performs very closely to the top two methods on the Diverse Verbs metric, indicating its ability to generate dynamic and diverse actions in narratives.

It is also worth mentioning that the diversity metrics of Qwen2-72B-Instruct are the lowest, which is consistent with the previously observed phenomenon of repeated paragraphs in its outputs.GPT-4o performs moderately well on DistinctL-2 and DistinctL-5, with a noticeable lead only on Diverse Verbs.

These quantitative metrics demonstrate that \method{} exhibits significant textual diversity in story generation. Textual diversity consititutes a crucial dimension for evaluating story quality, as it enhances the novelty and appeal of narrative content while preventing textual repetition and monotony, thereby improving the reader experience.

\section{Conclusion}

In this paper. we propose a novel approach called \method{}, which enhances the coherence and engagement of automatically generated stories through improved plot-planning mechanisms. Addressing issues such as structural inconsistency and narrative disfluency in long-form content, \method{} ensures more controlled and logical development of key events and character arcs by managing plot nodes and enabling dynamic interactions between different parts of the story. By leveraging flexible interactions between the \storyline{} and the narrative entity knowledge graph (\nekg), \method{} achieves a more cohesive narrative flow.

Experiments show that \method{} outperforms state-of-the-art models in generating more engaging and coherent stories. Future work will focus on handling more complex narrative structures and optimizing performance across various genres, unlocking new possibilities for storytelling in interactive systems and creative industries.

\section*{Limitations}

Despite generating relatively high-quality stories, a substantial gap remains between these outputs and the ultimate goal of producing a complete novel. The aspiration is to create works that fully embody the depth, complexity, and narrative richness inherent in the term "novel".

Currently, the field of story generation lacks specialized evaluation metrics capable of providing a comprehensive assessment of generated narratives. This absence represents a significant challenge, hindering the accurate evaluation and comparison of different story generation models.

\section*{Ethical Statements}

This study employs LLMs to generate and evaluate story content based on prompts. Although no personal data or animal subjects are involved, ethical risks such as copyright infringement or unauthorized reuse of intellectual property must be considered. LLMs are trained on large, publicly available datasets with unknown sources, which may lead to inadvertent replication of copyrighted material. To mitigate this, we review generated content to ensure originality and adherence to fair use principles.

The evaluation process uses automated LLM-based assessments to reduce subjective bias. However, we acknowledge the limitations of such tools and incorporate multi-dimensional metrics and careful judgment in our analysis.

\section*{Acknowledgements}

Min Yang was supported by National Key Research and Development Program of China (2022YFF0902100), National Natural Science Foundation of China (Grant No. 62376262), the Natural Science Foundation of Guangdong Province of China (2024A1515030166, 2025B1515020032), Shenzhen Science and Technology Innovation Program (KQTD20190929172835662).

\bibliography{custom}

\begin{thebibliography}{44}
\providecommand{\natexlab}[1]{#1}

\bibitem[{Achiam et~al.(2023)Achiam, Adler, Agarwal, Ahmad, Akkaya, Aleman, Almeida, Altenschmidt, Altman, Anadkat et~al.}]{achiam2023gpt}
Josh Achiam, Steven Adler, Sandhini Agarwal, Lama Ahmad, Ilge Akkaya, Florencia~Leoni Aleman, Diogo Almeida, Janko Altenschmidt, Sam Altman, Shyamal Anadkat, et~al. 2023.
\newblock Gpt-4 technical report.
\newblock \emph{arXiv preprint arXiv:2303.08774}.

\bibitem[{AI@Meta(2024)}]{llama3modelcard}
AI@Meta. 2024.
\newblock \href {https://github.com/meta-llama/llama3/blob/main/MODEL_CARD.md} {Llama 3 model card}.

\bibitem[{Alabdulkarim et~al.(2021)Alabdulkarim, Li, and Peng}]{alabdulkarim2021automatic}
Amal Alabdulkarim, Siyan Li, and Xiangyu Peng. 2021.
\newblock Automatic story generation: Challenges and attempts.
\newblock \emph{arXiv preprint arXiv:2102.12634}.

\bibitem[{Alhussain and Azmi(2021)}]{alhussain2021automatic}
Arwa~I Alhussain and Aqil~M Azmi. 2021.
\newblock Automatic story generation: A survey of approaches.
\newblock \emph{ACM Computing Surveys (CSUR)}, 54(5):1--38.

\bibitem[{Bai et~al.(2023)Bai, Bai, Chu, Cui, Dang, Deng, Fan, Ge, Han, Huang, Hui, Ji, Li, Lin, Lin, Liu, Liu, Lu, Lu, Ma, Men, Ren, Ren, Tan, Tan, Tu, Wang, Wang, Wang, Wu, Xu, Xu, Yang, Yang, Yang, Yang, Yao, Yu, Yuan, Yuan, Zhang, Zhang, Zhang, Zhang, Zhou, Zhou, Zhou, and Zhu}]{qwen}
Jinze Bai, Shuai Bai, Yunfei Chu, Zeyu Cui, Kai Dang, Xiaodong Deng, Yang Fan, Wenbin Ge, Yu~Han, Fei Huang, Binyuan Hui, Luo Ji, Mei Li, Junyang Lin, Runji Lin, Dayiheng Liu, Gao Liu, Chengqiang Lu, Keming Lu, Jianxin Ma, Rui Men, Xingzhang Ren, Xuancheng Ren, Chuanqi Tan, Sinan Tan, Jianhong Tu, Peng Wang, Shijie Wang, Wei Wang, Shengguang Wu, Benfeng Xu, Jin Xu, An~Yang, Hao Yang, Jian Yang, Shusheng Yang, Yang Yao, Bowen Yu, Hongyi Yuan, Zheng Yuan, Jianwei Zhang, Xingxuan Zhang, Yichang Zhang, Zhenru Zhang, Chang Zhou, Jingren Zhou, Xiaohuan Zhou, and Tianhang Zhu. 2023.
\newblock Qwen technical report.
\newblock \emph{arXiv preprint arXiv:2309.16609}.

\bibitem[{Bai et~al.(2024)Bai, Zhang, Lv, Zheng, Zhu, Hou, Dong, Tang, and Li}]{bai2024longwriter}
Yushi Bai, Jiajie Zhang, Xin Lv, Linzhi Zheng, Siqi Zhu, Lei Hou, Yuxiao Dong, Jie Tang, and Juanzi Li. 2024.
\newblock Longwriter: Unleashing 10,000+ word generation from long context llms.
\newblock \emph{arXiv preprint arXiv:2408.07055}.

\bibitem[{Brande(2013)}]{brande2013becoming}
Dorothea Brande. 2013.
\newblock from becoming a writer.
\newblock In \emph{Creative Writing}, pages 424--427. Routledge.

\bibitem[{Brown(2020)}]{brown2020language}
Tom~B Brown. 2020.
\newblock Language models are few-shot learners.
\newblock \emph{arXiv preprint arXiv:2005.14165}.

\bibitem[{Clark(2008)}]{clark2008writing}
Roy~Peter Clark. 2008.
\newblock \emph{Writing tools: 55 essential strategies for every writer}.
\newblock Little, Brown Spark.

\bibitem[{Cook(2011)}]{cook2011plotto}
William Cook. 2011.
\newblock \emph{PLOTTO: the master book of all plots}.
\newblock Tin House Books.

\bibitem[{Cupitt(1991)}]{Cupitt1991-CUPWIA}
Don Cupitt. 1991.
\newblock \emph{What is a Story?}
\newblock Trinity PressIntl.

\bibitem[{Devlin et~al.(2019)Devlin, Chang, Lee, and Toutanova}]{devlin-etal-2019-bert}
Jacob Devlin, Ming-Wei Chang, Kenton Lee, and Kristina Toutanova. 2019.
\newblock \href {https://doi.org/10.18653/v1/N19-1423} {{BERT}: Pre-training of deep bidirectional transformers for language understanding}.
\newblock In \emph{Proceedings of the 2019 Conference of the North {A}merican Chapter of the Association for Computational Linguistics: Human Language Technologies, Volume 1 (Long and Short Papers)}, pages 4171--4186, Minneapolis, Minnesota. Association for Computational Linguistics.

\bibitem[{Dong et~al.(2022)Dong, Li, Dai, Zheng, Wu, Chang, Sun, Xu, and Sui}]{dong2022survey}
Qingxiu Dong, Lei Li, Damai Dai, Ce~Zheng, Zhiyong Wu, Baobao Chang, Xu~Sun, Jingjing Xu, and Zhifang Sui. 2022.
\newblock A survey on in-context learning.
\newblock \emph{arXiv preprint arXiv:2301.00234}.

\bibitem[{Dubey et~al.(2024)Dubey, Jauhri, Pandey, Kadian, Al-Dahle, Letman, Mathur, Schelten, Yang, Fan et~al.}]{dubey2024llama}
Abhimanyu Dubey, Abhinav Jauhri, Abhinav Pandey, Abhishek Kadian, Ahmad Al-Dahle, Aiesha Letman, Akhil Mathur, Alan Schelten, Amy Yang, Angela Fan, et~al. 2024.
\newblock The llama 3 herd of models.
\newblock \emph{arXiv preprint arXiv:2407.21783}.

\bibitem[{Fan et~al.(2018)Fan, Lewis, and Dauphin}]{fan2018hierarchical}
Angela Fan, Mike Lewis, and Yann Dauphin. 2018.
\newblock Hierarchical neural story generation.
\newblock In \emph{Conference of the Association for Computational Linguistics (ACL)}.

\bibitem[{Fan et~al.(2019)Fan, Lewis, and Dauphin}]{fan-etal-2019-strategies}
Angela Fan, Mike Lewis, and Yann Dauphin. 2019.
\newblock \href {https://doi.org/10.18653/v1/P19-1254} {Strategies for structuring story generation}.
\newblock In \emph{Proceedings of the 57th Annual Meeting of the Association for Computational Linguistics}, pages 2650--2660, Florence, Italy. Association for Computational Linguistics.

\bibitem[{GLM et~al.(2024)GLM, Zeng, Xu, Wang, Zhang, Yin, Rojas, Feng, Zhao, Lai, Yu, Wang, Sun, Zhang, Cheng, Gui, Tang, Zhang, Li, Zhao, Wu, Zhong, Liu, Huang, Zhang, Zheng, Lu, Duan, Zhang, Cao, Yang, Tam, Zhao, Liu, Xia, Zhang, Gu, Lv, Liu, Liu, Yang, Song, Zhang, An, Xu, Niu, Yang, Li, Bai, Dong, Qi, Wang, Yang, Du, Hou, and Wang}]{glm2024chatglm}
Team GLM, Aohan Zeng, Bin Xu, Bowen Wang, Chenhui Zhang, Da~Yin, Diego Rojas, Guanyu Feng, Hanlin Zhao, Hanyu Lai, Hao Yu, Hongning Wang, Jiadai Sun, Jiajie Zhang, Jiale Cheng, Jiayi Gui, Jie Tang, Jing Zhang, Juanzi Li, Lei Zhao, Lindong Wu, Lucen Zhong, Mingdao Liu, Minlie Huang, Peng Zhang, Qinkai Zheng, Rui Lu, Shuaiqi Duan, Shudan Zhang, Shulin Cao, Shuxun Yang, Weng~Lam Tam, Wenyi Zhao, Xiao Liu, Xiao Xia, Xiaohan Zhang, Xiaotao Gu, Xin Lv, Xinghan Liu, Xinyi Liu, Xinyue Yang, Xixuan Song, Xunkai Zhang, Yifan An, Yifan Xu, Yilin Niu, Yuantao Yang, Yueyan Li, Yushi Bai, Yuxiao Dong, Zehan Qi, Zhaoyu Wang, Zhen Yang, Zhengxiao Du, Zhenyu Hou, and Zihan Wang. 2024.
\newblock \href {https://arxiv.org/abs/2406.12793} {Chatglm: A family of large language models from glm-130b to glm-4 all tools}.
\newblock \emph{Preprint}, arXiv:2406.12793.

\bibitem[{Guan et~al.(2020)Guan, Huang, Zhao, Zhu, and Huang}]{guan2020knowledge}
Jian Guan, Fei Huang, Zhihao Zhao, Xiaoyan Zhu, and Minlie Huang. 2020.
\newblock A knowledge-enhanced pretraining model for commonsense story generation.
\newblock \emph{Transactions of the Association for Computational Linguistics}, 8:93--108.

\bibitem[{Kwon et~al.(2023)Kwon, Li, Zhuang, Sheng, Zheng, Yu, Gonzalez, Zhang, and Stoica}]{kwon2023efficient}
Woosuk Kwon, Zhuohan Li, Siyuan Zhuang, Ying Sheng, Lianmin Zheng, Cody~Hao Yu, Joseph~E. Gonzalez, Hao Zhang, and Ion Stoica. 2023.
\newblock Efficient memory management for large language model serving with pagedattention.
\newblock In \emph{Proceedings of the ACM SIGOPS 29th Symposium on Operating Systems Principles}.

\bibitem[{Lamott(1995)}]{lamott1995bird}
Anne Lamott. 1995.
\newblock \emph{Bird by bird: Some instructions on writing and life}.
\newblock Vintage.

\bibitem[{Levine et~al.(2022)Levine, Wies, Jannai, Navon, Hoshen, and Shashua}]{levine2022the}
Yoav Levine, Noam Wies, Daniel Jannai, Dan Navon, Yedid Hoshen, and Amnon Shashua. 2022.
\newblock \href {https://openreview.net/forum?id=lnEaqbTJIRz} {The inductive bias of in-context learning: Rethinking pretraining example design}.
\newblock In \emph{International Conference on Learning Representations}.

\bibitem[{Li et~al.(2013)Li, Lee-Urban, Johnston, and Riedl}]{li2013story}
Boyang Li, Stephen Lee-Urban, George Johnston, and Mark Riedl. 2013.
\newblock Story generation with crowdsourced plot graphs.
\newblock In \emph{Proceedings of the AAAI Conference on Artificial Intelligence}, volume~27, pages 598--604.

\bibitem[{Li et~al.(2016)Li, Galley, Brockett, Gao, and Dolan}]{li-etal-2016-diversity}
Jiwei Li, Michel Galley, Chris Brockett, Jianfeng Gao, and Bill Dolan. 2016.
\newblock \href {https://doi.org/10.18653/v1/N16-1014} {A diversity-promoting objective function for neural conversation models}.
\newblock In \emph{Proceedings of the 2016 Conference of the North {A}merican Chapter of the Association for Computational Linguistics: Human Language Technologies}, pages 110--119, San Diego, California. Association for Computational Linguistics.

\bibitem[{Li et~al.(2024)Li, Chiang, Frick, Dunlap, Wu, Zhu, Gonzalez, and Stoica}]{li2024crowdsourced}
Tianle Li, Wei-Lin Chiang, Evan Frick, Lisa Dunlap, Tianhao Wu, Banghua Zhu, Joseph~E Gonzalez, and Ion Stoica. 2024.
\newblock From crowdsourced data to high-quality benchmarks: Arena-hard and benchbuilder pipeline.
\newblock \emph{arXiv preprint arXiv:2406.11939}.

\bibitem[{Liu et~al.(2023)Liu, Yuan, Fu, Jiang, Hayashi, and Neubig}]{liu2023pre}
Pengfei Liu, Weizhe Yuan, Jinlan Fu, Zhengbao Jiang, Hiroaki Hayashi, and Graham Neubig. 2023.
\newblock Pre-train, prompt, and predict: A systematic survey of prompting methods in natural language processing.
\newblock \emph{ACM Computing Surveys}, 55(9):1--35.

\bibitem[{Meehan(1977)}]{meehan1977tale}
James~R Meehan. 1977.
\newblock Tale-spin, an interactive program that writes stories.
\newblock In \emph{Ijcai}, volume~77, pages 91--98.

\bibitem[{Mirowski et~al.(2023)Mirowski, Mathewson, Pittman, and Evans}]{mirowski2023co}
Piotr Mirowski, Kory~W Mathewson, Jaylen Pittman, and Richard Evans. 2023.
\newblock Co-writing screenplays and theatre scripts with language models: Evaluation by industry professionals.
\newblock In \emph{Proceedings of the 2023 CHI Conference on Human Factors in Computing Systems}, pages 1--34.

\bibitem[{OpenAI()}]{HelloGPT4o}
OpenAI.
\newblock Hello gpt-4o | openai.
\newblock \url{https://openai.com/index/hello-gpt-4o/}.

\bibitem[{Peng et~al.(2022)Peng, Li, Wiegreffe, and Riedl}]{peng-etal-2022-inferring}
Xiangyu Peng, Siyan Li, Sarah Wiegreffe, and Mark Riedl. 2022.
\newblock \href {https://doi.org/10.18653/v1/2022.findings-emnlp.520} {Inferring the reader: Guiding automated story generation with commonsense reasoning}.
\newblock In \emph{Findings of the Association for Computational Linguistics: EMNLP 2022}, pages 7008--7029, Abu Dhabi, United Arab Emirates. Association for Computational Linguistics.

\bibitem[{Peng et~al.(2021)Peng, Xie, Alabdulkarim, Kayam, Dani, and Riedl}]{peng2021guiding}
Xiangyu Peng, Kaige Xie, Amal Alabdulkarim, Harshith Kayam, Samihan Dani, and Mark~O Riedl. 2021.
\newblock Guiding neural story generation with reader models.
\newblock \emph{arXiv preprint arXiv:2112.08596}.

\bibitem[{Rashkin et~al.(2020)Rashkin, Celikyilmaz, Choi, and Gao}]{rashkin2020plotmachines}
Hannah Rashkin, Asli Celikyilmaz, Yejin Choi, and Jianfeng Gao. 2020.
\newblock Plotmachines: Outline-conditioned generation with dynamic plot state tracking.
\newblock \emph{arXiv preprint arXiv:2004.14967}.

\bibitem[{Tearse et~al.(2010)Tearse, Wardrip-Fruin, and Mateas}]{tearse2010minstrel}
Brandon Tearse, Noah Wardrip-Fruin, and Michael Mateas. 2010.
\newblock Minstrel remixed: Procedurally generating stories.
\newblock In \emph{Proceedings of the AAAI Conference on Artificial Intelligence and Interactive Digital Entertainment}, volume~6, pages 192--197.

\bibitem[{Touvron et~al.(2023)Touvron, Lavril, Izacard, Martinet, Lachaux, Lacroix, Rozi{\`e}re, Goyal, Hambro, Azhar et~al.}]{touvron2023llama}
Hugo Touvron, Thibaut Lavril, Gautier Izacard, Xavier Martinet, Marie-Anne Lachaux, Timoth{\'e}e Lacroix, Baptiste Rozi{\`e}re, Naman Goyal, Eric Hambro, Faisal Azhar, et~al. 2023.
\newblock Llama: Open and efficient foundation language models.
\newblock \emph{arXiv preprint arXiv:2302.13971}.

\bibitem[{Turner(1993)}]{turner1993minstrel}
Scott~R Turner. 1993.
\newblock \emph{Minstrel: a computer model of creativity and storytelling}.
\newblock University of California, Los Angeles.

\bibitem[{Vaswani et~al.(2017)Vaswani, Shazeer, Parmar, Uszkoreit, Jones, Gomez, Kaiser, and Polosukhin}]{vaswani2017attention}
Ashish Vaswani, Noam Shazeer, Niki Parmar, Jakob Uszkoreit, Llion Jones, Aidan~N Gomez, {\L}ukasz Kaiser, and Illia Polosukhin. 2017.
\newblock Attention is all you need.
\newblock \emph{Advances in neural information processing systems}, 30.

\bibitem[{Wang et~al.(2023)Wang, Lin, Yu, Hu, and Karlsson}]{wang2023open}
Yuxin Wang, Jieru Lin, Zhiwei Yu, Wei Hu, and B{\"o}rje~F Karlsson. 2023.
\newblock Open-world story generation with structured knowledge enhancement: A comprehensive survey.
\newblock \emph{Neurocomputing}, page 126792.

\bibitem[{Wei et~al.(2022)Wei, Tay, Bommasani, Raffel, Zoph, Borgeaud, Yogatama, Bosma, Zhou, Metzler, Chi, Hashimoto, Vinyals, Liang, Dean, and Fedus}]{wei2022emergent}
Jason Wei, Yi~Tay, Rishi Bommasani, Colin Raffel, Barret Zoph, Sebastian Borgeaud, Dani Yogatama, Maarten Bosma, Denny Zhou, Donald Metzler, Ed~H. Chi, Tatsunori Hashimoto, Oriol Vinyals, Percy Liang, Jeff Dean, and William Fedus. 2022.
\newblock \href {https://openreview.net/forum?id=yzkSU5zdwD} {Emergent abilities of large language models}.
\newblock \emph{Transactions on Machine Learning Research}.
\newblock Survey Certification.

\bibitem[{Yang et~al.(2022{\natexlab{a}})Yang, Klein, Peng, and Tian}]{yang2022doc}
Kevin Yang, Dan Klein, Nanyun Peng, and Yuandong Tian. 2022{\natexlab{a}}.
\newblock Doc: Improving long story coherence with detailed outline control.
\newblock \emph{arXiv preprint arXiv:2212.10077}.

\bibitem[{Yang et~al.(2022{\natexlab{b}})Yang, Tian, Peng, and Klein}]{yang2022re3}
Kevin Yang, Yuandong Tian, Nanyun Peng, and Dan Klein. 2022{\natexlab{b}}.
\newblock Re3: Generating longer stories with recursive reprompting and revision.
\newblock \emph{arXiv preprint arXiv:2210.06774}.

\bibitem[{Yao et~al.(2019)Yao, Peng, Weischedel, Knight, Zhao, and Yan}]{yao2019plan}
Lili Yao, Nanyun Peng, Ralph Weischedel, Kevin Knight, Dongyan Zhao, and Rui Yan. 2019.
\newblock Plan-and-write: Towards better automatic storytelling.
\newblock In \emph{Proceedings of the AAAI Conference on Artificial Intelligence}, volume~33, pages 7378--7385.

\bibitem[{Zhang et~al.(2023)Zhang, Li, Liu, yang, Liu, and Yang}]{zhang2023marathon}
Lei Zhang, Yunshui Li, Ziqiang Liu, Jiaxi yang, Junhao Liu, and Min Yang. 2023.
\newblock \href {https://arxiv.org/abs/2312.09542} {Marathon: A race through the realm of long context with large language models}.
\newblock \emph{Preprint}, arXiv:2312.09542.

\bibitem[{Zhang et~al.(2022)Zhang, Roller, Goyal, Artetxe, Chen, Chen, Dewan, Diab, Li, Lin et~al.}]{zhang2022opt}
Susan Zhang, Stephen Roller, Naman Goyal, Mikel Artetxe, Moya Chen, Shuohui Chen, Christopher Dewan, Mona Diab, Xian Li, Xi~Victoria Lin, et~al. 2022.
\newblock Opt: Open pre-trained transformer language models.
\newblock \emph{arXiv preprint arXiv:2205.01068}.

\bibitem[{Zhao et~al.(2021)Zhao, Wallace, Feng, Klein, and Singh}]{zhao2021calibrate}
Zihao Zhao, Eric Wallace, Shi Feng, Dan Klein, and Sameer Singh. 2021.
\newblock Calibrate before use: Improving few-shot performance of language models.
\newblock In \emph{International conference on machine learning}, pages 12697--12706. PMLR.

\bibitem[{Zhou et~al.(2024)Zhou, Liu, Xu, Iyer, Sun, Mao, Ma, Efrat, Yu, Yu et~al.}]{zhou2024lima}
Chunting Zhou, Pengfei Liu, Puxin Xu, Srinivasan Iyer, Jiao Sun, Yuning Mao, Xuezhe Ma, Avia Efrat, Ping Yu, Lili Yu, et~al. 2024.
\newblock Lima: Less is more for alignment.
\newblock \emph{Advances in Neural Information Processing Systems}, 36.

\end{thebibliography}

\appendix
\clearpage

\onecolumn
\section{\textsc{WritingPrompts} Dataset}
\label{appendix:WritingPrompts Dataset}

In this section, we detail the \textsc{WritingPrompts} dataset \citep{fan2018hierarchical}, including its tag system, structural composition, and key characteristics.

Prompt tags, enclosed within square brackets, help users organize or filter prompts by type. The dataset includes 11 Reddit-defined tags, with descriptions summarized in Table \ref{tab:WritingPrompts Tags}. For comprehensive tag definitions and usage guidelines refer to the official Reddit documentation\footnote{\url{https://www.reddit.com/r/WritingPrompts/wiki/how\_to\_tag\_prompts}}.

\begin{table}[H]
	\centering
	\begin{tabular}{llp{9cm}}
		\toprule
		\textbf{Tag}    & \textbf{Full Name}   & \textbf{Description}                                                                                                                                                               \\
		\midrule
		\texttt{[ WP ]} & Writing Prompt       & Basic prompts with no restrictions, providing simple ideas to inspire narrative fiction, poetry, or other forms of writing.                                                        \\
		\midrule
		\texttt{[ SP ]} & Simple Prompt        & A basic prompt with a title containing no more than 100 characters.                                                                                                                \\
		\midrule
		\texttt{[ EU ]} & Established Universe & For prompts set in pre-established fictional worlds, encouraging fan fiction that adds new scenarios to well-known series and characters.                                          \\
		\midrule
		\texttt{[ CW ]} & Constrained Writing  & Includes constraints or limitations, allowing restrictions on words, word limits, or specific writing styles.                                                                      \\
		\midrule
		\texttt{[ TT ]} & Theme Thursday       & Focus on weekly themed writing styles, requiring conformity to the specific theme of the week.                                                                                     \\
		\midrule
		\texttt{[ PM ]} & Prompt Me            & Encourage responding to comments as prompts, expanding creative output and writing skills by treating responses as prompts.                                                        \\
		\midrule
		\texttt{[ MP ]} & Media Prompt         & Use audio or visual media to inspire writing, offering an alternative to traditional text-based prompts.                                                                           \\
		\midrule
		\texttt{[ IP ]} & Image Prompt         & Linked images to inspire stories or poems, similar to Media Prompts, but AI-generated art is not allowed.                                                                          \\
		\midrule
		\texttt{[ PI ]} & Prompt Inspired      & Consists of standalone responses to prompts at least three days old, with the story posted in one post or in comments if too long, and must include a link to the original prompt. \\
		\midrule
		\texttt{[ OT ]} & Off Topic            & Off topic and not prompts, but are writing related. Not for stories.                                                                                                               \\
		\midrule
		\texttt{[ RF ]} & Reality Fiction      & Focus on events that have occurred or could happen in the real world to unknown people, excluding future possibilities.                                                            \\
		\bottomrule
	\end{tabular}
	\caption{\label{tab:WritingPrompts Tags}
		Prompt tags  in \textsc{WritingPrompts} and their descriptions.
	}
\end{table}

\section{Details Of The Formula Derivation}
\label{appendix:appDetails Of The Formula Derivation}
In this section, we provide the details of the formula derivation.

\begin{tcolorbox}[title=\textit{Pair Rank Brier Score Derivation},
    colback=lightblue!20!white,  
    colframe=lightgray!30!black,  
    breakable,
    pad at break=3mm]

Bootstrapping is a statistical technique used to estimate the distribution of an estimator by sampling with replacement from the original dataset. This method is widely used for constructing confidence intervals, particularly in tasks such as LLM leaderboards. For metrics like the \textbf{Pairwise Rank Brier Score}, estimating the probability distribution of rank-based model performance is essential.

\subsection*{Problem Setting}

Consider a benchmark dataset $D = \{x_1, x_2, \dots, x_{|D|}\}$ and a scoring function $f$ that evaluates the performance of $n$ models $\pi_1, \pi_2, \dots, \pi_n$ on this dataset. Let $D^*$ denote a bootstrap sample of $D$, and let $f^*(\pi_i, D^*)$ represent the bootstrapped performance score for model $\pi_i$ using the dataset $D^*$. For simplicity, we denote this as $f^*(\pi_i)$.

\subsection*{Step 1: Probabilistic Prediction}

To evaluate the accuracy of the benchmark's probabilistic predictions, we compute the probability that model $\pi_i$ performs worse than model $\pi_j$:
\begin{equation}
    P(f^*(\pi_i) < f^*(\pi_j)).
\end{equation}

\subsection*{Step 2: Distribution Assumption}

The bootstrapped scores $f^*(\pi_i)$ and $f^*(\pi_j)$ follow an empirical distribution. In many cases, these scores asymptotically converge to a normal distribution due to the Central Limit Theorem (CLT):
\begin{equation}
    f^*(\pi_i) \sim \mathcal{N}(\mu_i, \sigma_i^2), \quad f^*(\pi_j) \sim \mathcal{N}(\mu_j, \sigma_j^2),
\end{equation}
where $\mu_i$ and $\sigma_i^2$ are the bootstrapped mean and variance for model $\pi_i$, respectively.

When the normality assumption holds, the probability $P(f^*(\pi_i) < f^*(\pi_j))$ can be computed as:
\begin{equation}
    P(f^*(\pi_i) < f^*(\pi_j)) = \Phi\left(\frac{\mu_j - \mu_i}{\sqrt{\sigma_i^2 + \sigma_j^2}}\right),
\end{equation}
where $\Phi(\cdot)$ is the cumulative distribution function (CDF) of the standard normal distribution.

If the normality assumption does not hold, the probability can still be estimated empirically from the distribution of the bootstrapped scores.

\subsection*{Step 3: Ground Truth Outcome}

Define the ground truth outcome for the model pair $(\pi_i, \pi_j)$ as:
\begin{equation}
    O_{\pi_i < \pi_j} = 
    \begin{cases} 
        1, & \text{if model $\pi_i$ performs worse than $\pi_j$ on the ground truth evaluation metric}, \\
        0, & \text{otherwise}.
    \end{cases}
\end{equation}

\subsection*{Step 4: Brier Score Loss}

The \textbf{Brier Score Loss} is calculated over the benchmark's probabilistic predictions for each model pair $(\pi_i, \pi_j)$ with respect to the ground truth outcome $O$:
\begin{equation}
    \text{Brier Score Loss} = \frac{1}{N} \sum_{i,j} \left(P(f^*(\pi_i) < f^*(\pi_j)) - O_{\pi_i < \pi_j}\right)^2,
\end{equation}
where $N$ is the total number of model pairs.

\end{tcolorbox}

\section{Prompt Templates}
\label{appendix:Prompt Templates}

In this section, we provide a comprehensive list of prompt templates used in this study.

\subsection{Chat Template}
\label{appendix:Chat Template}

For GPT-4o, LongWriter Models, Qwen2-72B-Instruct and Meta-Llama-3.1-70B-Instruct, we employ the following template for story generation, with an explicit requirement  for length: the story must exceed 3,000 words and prioritize maximal narrative expansion. This focus on extended narrative generation aligns with our objective of producing long-form stories.

\begin{prompt}{Chat Template}\label{Chat Template}
	You should write an engaging story based on the following requirements and writing prompt.
	\\
        \\
	\text{[Requirements] :}
        \\
        \\
	1. Feel free to use creativity to expand on the prompt and create an interesting and captivating narrative.
        \\
        \\
	2. Ensure the story is at least 3,000 words long, but strive for more if the story naturally allows for it.
        \\
        \\
	3. Create well-developed characters with distinct personalities, motivations, and backstories.
        \\
        \\
	4. The plot should have multiple layers—include subplots or side characters that enrich the main storyline and add complexity to the story.
        \\
        \\
	5. Balance the pacing of the narrative.
        \\
        \\
	6. Use rich and evocative language that engages all the senses—sight, sound, smell, touch, and taste—to create a fully immersive reading experience.
        \\
        \\
	7. Ensure the grammar, sentence structure, and overall coherence are polished, and the story flows smoothly from beginning to end.
        \\
        \\
	8. Return only the final, complete generated story.
	\\
        \\
	\text{[Writing Prompts]:\textcolor{blue}{\{WP\}}}

\end{prompt}

\subsection{Prospective Evaluation Template}
\label{appendix:Prospective Evaluation Template}
 Prospective evaluation prompt is derived by modifying the preference evaluation prompt, but we recognize that its inclusion is essential for reproducibility and clarity. When the aspect to be judged is specified, the corresponding \textbf{metric} and \textbf{checklist} in \textsc{SYSTEM\_PROMPT} will be replaced accordingly. 
\newpage

\begin{prompt}{Prospective Evaluation Template}\label{Prospective Evaluation Template}
SYSTEM\_PROMPT = (
{
\setlength{\parindent}{2em}

    "Please act as an impartial judge and evaluate the quality of the responses provided by two AI assistants to a user prompt. You will be given assistant A's answer (Story A) and assistant B's answer (Story B). Your job is to evaluate which assistant's story is better.\textbackslash n\textbackslash n"
    
    "When evaluating the two stories, you should consider the story must be at least 3,000 words long, and we strongly encourage expanding the story beyond this minimum if the narrative allows for it. You should focus on this factor: {\{\textcolor{blue}{metric}\}} \textbackslash n\textbackslash n" 
   
    "Here are the checklists of this facter:\textbackslash n"
    
    '\{\{"checklists": \{\textcolor{blue}{checklists}\}\}\}\textbackslash n\textbackslash n'
    
    "You should be strict but fair in your evaluation.\textbackslash n\textbackslash n"
    
    "After thinking your analysis and justification, you must output only one of the following choices as your final verdict with a label:\textbackslash n\textbackslash n"
    
    "1. Assistant A is significantly better: [[A>>B]]\textbackslash n"
    
    "2. Assistant A is slightly better: [[A>B]]\textbackslash n"
    
    "3. Tie, relatively the same: [[A=B]]\textbackslash n"
    
    "4. Assistant B is slightly better: [[B>A]]\textbackslash n"
    
    "5. Assistant B is significantly better: [[B>>A]]\textbackslash n\textbackslash n"
    
    "Just a reminder, you only need to respond in the following format,do not return any unrelated information:"
    
    'Example output: "My final verdict is tie: [[A=B]]".'
}    

)
    \\
    \\
CHECKLIST = \{

\setlength{\parindent}{2em}
"creativity": [
\setlength{\parindent}{4em}
        
        "Uniqueness of the plot setting: Does the story's background, time, location, or plot present a distinctive setting that breaks away from conventional patterns?",
        
        "Innovation in character design: Do the characters have distinct traits, unconventional identities, or surprising personalities? Is there unexpected character development or interaction?",
        
        "Creativity in narrative technique: Does the story employ novel narrative structures or techniques, such as nonlinear storytelling, multiple perspectives, or unconventional modes of expression?",
        
        "Fresh interpretation of common themes: Even if the story uses common themes (such as love, adventure, or conflict), does it offer a fresh perspective, different emotional layers, or unique interpretations?",
        
        "Incorporation of creative elements: Does the story include unexpected elements like fantasy, science fiction, or other imaginative concepts? Do these elements effectively enhance the story's appeal and uniqueness?",
        
    ],
    
\setlength{\parindent}{2em}
    "coherence": [
\setlength{\parindent}{4em}
        
        "Logical consistency of the plot: Do the events and plot developments in the story have clear cause-and-effect relationships? Are there any sudden or unreasonable twists?",
        
        "Coherence of the story structure: Does the story have a clear overall framework from beginning to end, with a natural progression of events? Does it avoid disjointed or erratic narration?",
        
        "Reasonableness of character behavior: Do the characters' actions and decisions align with their personality and background? Are the characters' responses in different situations consistent?",
        
        "Smoothness of the timeline: Is the progression of time in the story clear and coherent? Does the timeline remain consistent, avoiding confusing time jumps or unreasonable time gaps?",
        
        "Consistency of narrative tone: Is the language style and narrative tone of the story consistent throughout? Does it avoid unnecessary shifts in tone or abrupt changes in expression?",
        
    ],
    
\setlength{\parindent}{2em}
"engagement": [
\setlength{\parindent}{4em}
        
        "Gripping opening: Does the story capture attention from the very beginning? Is the introduction intriguing enough to make the reader want to continue?",
        
        "Sustained interest: Does the story maintain the reader's interest throughout? Are there moments of tension, excitement, or emotional depth that keep the reader engaged?",
        
        "Emotional connection: Does the story evoke any emotional response? Do the characters and their struggles create empathy or connection with the reader?",
        
        "Immersiveness of the narrative: Does the story create a vivid and immersive experience? Are the descriptions, dialogues, and world-building elements compelling and engaging?",
        
        "Pacing of the story: Is the pacing of the story appropriate? Does it avoid being too slow or too rushed, allowing the reader to stay immersed without losing interest?"
        
    ],
    
\setlength{\parindent}{2em}
"relevance": [
\setlength{\parindent}{4em}
        
        "Alignment with the theme: Does the story closely follow the given theme or prompt? Is the content aligned with the prompt's requirements, avoiding any unrelated or off-topic elements?",
        
        "Relevance of the plot to the task objective: Does the plot development revolve around the task or objective? Does it effectively convey the core message that the story is meant to deliver?",
        
        "Consistency of characters and setting with the task: Are the characters and setting in the story consistent with the task requirements or thematic setup? Does it avoid inappropriate or irrelevant elements?",
        
        "Match between story style and context: Is the style and tone of the story suitable for the given task or scenario? Does it maintain consistency without sudden shifts in style?",
        
        "Relevance of details and scenes: Are the details and scenes in the story relevant to the overall plot? Do they support and enhance the expression of the theme, avoiding unrelated or redundant descriptions?"
        
    ]
\\
\}

\end{prompt}

\subsection{Human Evaluation System}
\label{appendix:Human Evaluation System}
To address potential biases and limitations associated with solely relying on GPT-4o as an evaluator, we conducted human evaluation.

\begin{tcolorbox}[title=\textit{HumanEval Template},
    colback=mylightblue!50!white,
    colframe=mydarkblue!60!white,
    breakable,
    pad at break=3mm]
    
    \includegraphics[width=1\textwidth,page=1]{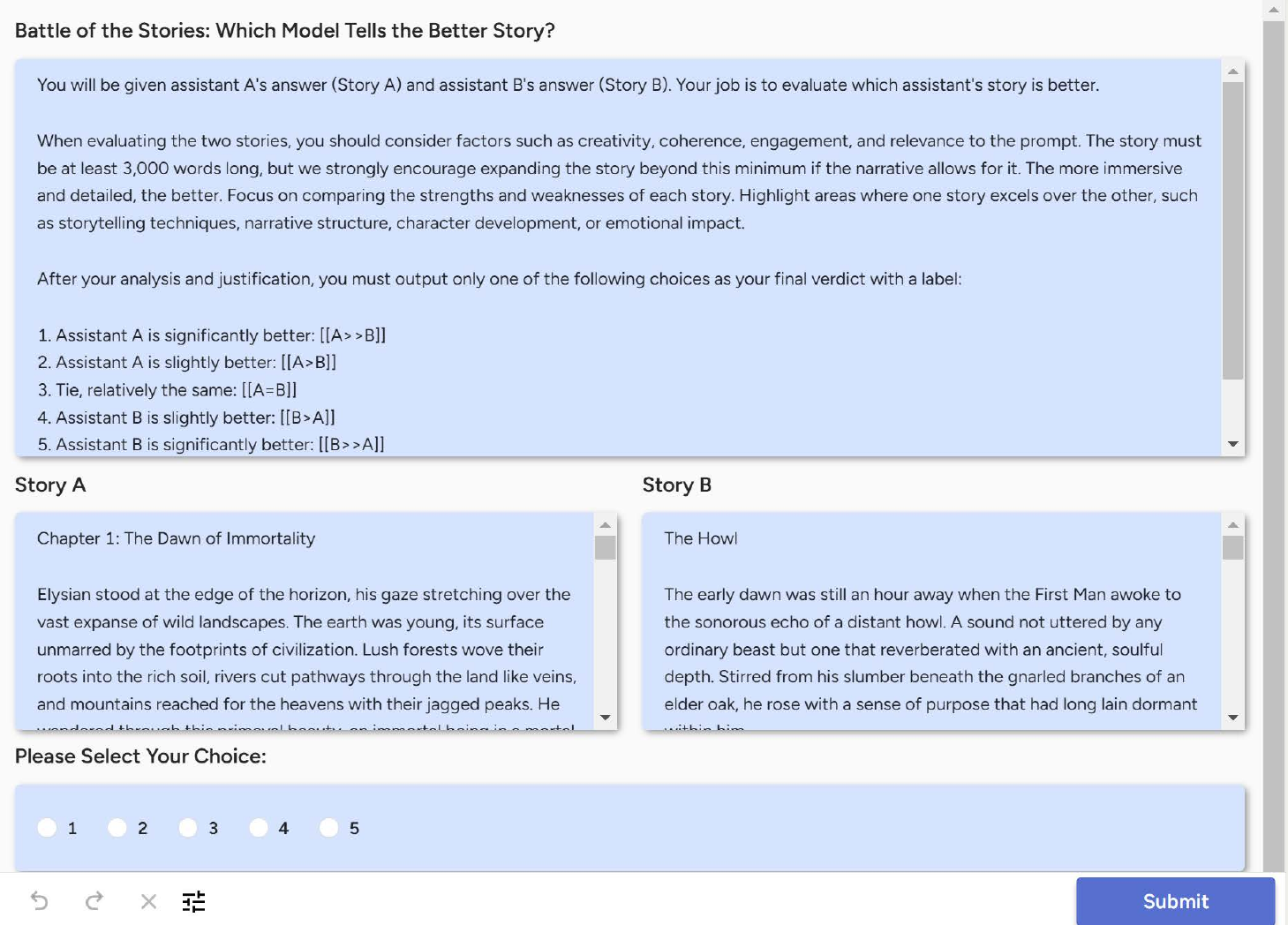}
    
\end{tcolorbox}

\section{CPN' Review Display}
\label{appendix:CPN' Review Display}
In the process of reviewing CPN, we use CPN\_Check to achieve this.In this section, we present the classification of CPN\_Check and provide 3 examples of CPN\_Check generated during the story generation process.

\begin{table}[H]
	\centering
	\begin{tabular}{lp{10cm}} 
		\toprule
		\textbf{Type} & \textbf{Description} \\ 
		\midrule
		\texttt{[ Theme Enhancement  ]} & Explore deeper philosophical meanings and core themes through plot or dialogue. \\
            \midrule
		\texttt{[ Logic Adjustment ]} & Optimize the causal relationships and pacing of the plot to ensure smooth and natural story progression. \\
            \midrule
		\texttt{[ Emotion Deepening ]} & Show authentic emotional changes and growth in characters to make them more vivid and relatable. \\
            \midrule
		\texttt{[ Mystery Enhancement ]} & Create a sense of mystery and the supernatural through detailed descriptions and suspenseful setups. \\
            \midrule
		\texttt{[ Plot Resolution ]} & Craft compelling chapter endings that both summarize the current plot and set up future developments. \\
            \midrule
		\texttt{[ Language Optimization ]} & Polish the narrative to make the language more vivid, elegant, and aligned with the story’s tone. \\
            \midrule
		\texttt{[ Redundancy Optimization ]} & Ensure consistency in timelines, settings, and character behavior while removing redundant or repetitive information. \\
		\bottomrule
	\end{tabular}
	\caption{\label{tab:CPN_Check's type and description}
		CPN\_Check's type and description
	}
\end{table}

\begin{tcolorbox}[title=\textit{Theme Enhancement },
    colback=pink!20!white,
    colframe=lightgray!50!black,
    breakable,
    pad at break=3mm]

The original plot node \{'subject': 'Ethan and Lily', 'verb': 'walk', 'object': 'shoreline'\} is redundant because it repeats the action of walking towards the shoreline, which has already been covered by the initial plot node \{'subject': 'Ethan', 'verb': 'walk', 'object': ''\}. By modifying the node to \{'subject': 'Ethan and Lily', 'verb': 'continue', 'object': 'walk'\}, it enriches the narrative by emphasizing their shared path and forwards the storyline towards forming a connection, fulfilling requirement 1 without unnecessary repetition.

\end{tcolorbox}

\begin{tcolorbox}[title=\textit{Logic Adjustment},
    colback=pink!20!white,
    colframe=lightgray!50!black,
    breakable,
    pad at break=3mm]

The original plot node 'Ethan speak Lily' is missing an event that precedes a conversation and is too abrupt. To enrich the main content and logically progress towards the connection they form, the node is modified to 'Ethan approach Lily'. This allows the natural development of their interaction as indicated in the abstract, where their exchange is tentative at first.

\end{tcolorbox}
\begin{tcolorbox}[title=\textit{Theme Enhancement},
    colback=pink!20!white,
    colframe=lightgray!50!black,
    breakable,
    pad at break=3mm]

The original node 'Ethan reflect on his intentions during the war' is somewhat redundant with previous nodes where Ethan acknowledges and confronts his past actions. By focusing on 'ethical dilemmas,' it aligns better with the chapter's theme of moral complexity and introspection, moving towards the concluding node of initiating introspection.

\end{tcolorbox}

\section{Illustrative Examples of Story Generation}
\label{appendix:Illustrative Examples of Story Generation}

In this section, we will present a number of story examples generated by \method{}. 
\\

The story is a poignant exploration of grief, healing, and the power of community, anchored by Elias Merrick's emotional journey as a war veteran returning to his scarred hometown. The narrative excels in its vivid descriptions and emotional depth, effectively immersing readers in Elias’s internal struggles and the postwar desolation of the town. The writing is reflective and evocative, with a balance of somber tones and hopeful moments. At its core, the story emphasizes resilience, human connection, and the potential for renewal, delivering a universal message about finding purpose and rebuilding amidst loss and trauma.

\label{appendix:Story Exposition}

\begin{tcolorbox}[
    title=\textit{Story Exposition},
    colback=brown!5!white,           
    colframe=brown!30!white,         
    breakable,
    pad at break=3mm
]

\textbf{Instruction:}
\,

Tell the story of a young man returning from a terrible war.

\tcblower
\textbf{Output:}

\begin{spacing}{1.5} 
{
\setlength{\parindent}{2em}

\small  

\textbf{chapter 1: Return to Shadows}

The hiss of the train brakes sliced through the stillness of the late afternoon, echoing off the worn stone of the platform. Elias Merrick stepped off the train, his polished boots clicking against the cracked concrete. The air here was different—heavier, almost leaden, carrying the scent of damp earth and decay with a faint metallic bitterness. His grip tightened around the handle of his suitcase, leather gloves creaking in protest. This was home, he reminded himself, though the word felt foreign, distant. 

The town stretched before him, sullen and subdued, its streets veiled in an uneasy silence. The once-bustling main square was nearly deserted. Window shutters hung askew, their faded paint peeling like dry bark; shop signs swung in halfhearted arcs, their chains moaning softly in the wind. The battered remains of a broken cart lay abandoned by the curb—a mute testament to a time when this place vibrated with life. As Elias walked, he noticed the war’s toll etched into every corner, as if the town itself had weathered battles alongside its people. Cracks spider-webbed through the facades of buildings he once recognized, and weeds burst defiantly through cobblestones, reclaiming terrain left untended. 

He passed the old general store and froze. The face of Mr. Graham, weary and drawn, appeared in the dim light of the shop’s window. The elderly shopkeeper opened his door as Elias approached, his expression caught somewhere between recognition and uncertainty. "Elias?" Mr. Graham said, his voice raspy and unsure, as though the name carried a weight he hadn’t prepared for. 

Elias tipped the brim of his hat in a stiff greeting. "Mr. Graham," he replied, his voice steady though his insides lurched awkwardly. The shopkeeper searched his face with furrowed brows, his gaze lingering too long, as if trying to reconcile the man before him with the boy he once knew. After a beat of silence, Mr. Graham nodded faintly, retreating back indoors with no further words. Elias walked on, his chest tightening. Did he feel pity in Mr. Graham’s hesitation? Or fear? He wasn’t sure which was worse.

A tug of familiarity guided his steps. He glanced at familiar landmarks—the clocktower missing half its face, the rusted fountain where children once clamored in summer heat. The cracks and neglect felt like bruises on memories he hadn’t yet decided to protect or abandon. Passing the childhood park, a rush of warmth and sorrow filled him. The swings hung silent now, blackened ropes gnarled with mildew, their arcs frozen in a time long gone. He paused, his hand curling around the rusted iron gate, his mind drifting back to a version of himself who felt this town might never change. That boy seemed like a stranger now.

As shades of dusk began to seep into the edges of the sky, Elias’s steps brought him closer to the long, weathered pathway to his family home. The sight of it made him hesitate, his boots planting firmly at the threshold of the yard. For a moment, a nervous energy churned in his chest. What would he find within these walls? He could almost see his younger self storming out the door years ago, suitcase in hand, eager to seize a world beyond these fences. He had been restless then. Now, he merely felt worn. 

The scrape of his boot against a loose pebble startled him into motion. He approached slowly, his fingers tracing the grooves of an old bench on the porch where he had spent countless summer evenings. The wood sagged under the weight of years, its edges rounded by time. He stumbled slightly as he leaned against it. Nearby, a lamppost leaned precariously, its surface etched with carvings—faded initials of friendships and love stories forgotten. His hand brushed against one, tracing its weathered lines. It was strange, he thought, to feel like the world had kept moving without him, only to find it had not moved at all.

As the last rays of sunlight dipped beneath the horizon, the melancholy of his hometown deepened. The alleyways, dimly lit, carried whispers of footfalls long since disappeared. Every shadow seemed to stretch with the weight of unspoken stories. His grip on the suitcase tightened as the sense of isolation among familiar streets settled heavily onto his shoulders. He let out a long breath, his sigh mingling with the cool, sharp night air. The faint smell of rain hung in the clouds above as he passed the darkened windows of his neighbors’ homes. He wondered where the laughter and warmth had gone, those vibrant echoes of life now resigned to silence.

When he reached the stone bridge overlooking the shallow river, he paused once more. The chill of the weathered railing seeped into his hands, grounding him. The water below mirrored the muted sky, broken only by the ripples of the occasional drifting leaf. He lingered there for long moments, staring into the distance, where the cobblestone path meandered toward his home. Somewhere above, a flock of birds wheeled in disjointed patterns before vanishing into the twilight.

The pathway stretched ahead, dusted with fallen leaves. It led to a house whose silhouette had haunted his dreams for years—a house that both welcomed and condemned. He stepped carefully across the pathway, his steps muffled by the debris of seasons left ungathered. Whatever awaited him inside, he had no choice but to meet it.

The door creaked slightly as he pushed it open. His mother greeted him with a hurried embrace, her hands trembling as they brushed over his shoulders and cheeks, as though reassuring herself that he was physically whole. She smiled, but her lips quivered at the edges, betraying the tears she fought to hold back. "You’ll stay for dinner," she said brightly, her tone fragile as spun glass. 

Behind her, his father stood by the doorway to the study. His handshake was firm, his grip briefly tightening as if assessing more than greeting. "Welcome back," he said gruffly before stepping aside. No words of comfort, no questions about the interim years, just a simple acknowledgment of return. Yet Elias caught the faint shift in his father’s expression, a fleeting vulnerability that flickered and vanished before it could take root. 

Retreating to his old bedroom later that night, Elias set down his suitcase and gazed at the room that had once been a sanctuary. The cracks in the ceiling traced new constellations amidst the familiar plaster. The bed creaked in a way that was both familiar and foreign. As he lay there staring into the dark, memories from the war seeped into the quiet, unbidden and vivid. The heat of explosions, the cries of distant commands, the hollowed faces of friends he had failed to save—they invaded the stillness. He breathed deeply, trying to tether himself to the present, to the shallow comfort of his neighborhood’s quiet, but even that felt alien. Outside, the distant hoot of an owl echoed faintly, a reminder that the night was long, and peace was as tenuous as the silence.

For a moment, he closed his eyes and thought of the boy who had once stared at this same ceiling, dreaming of far-off places, longing for anywhere but here. Now, here was all he had.

The morning sun filtered through a veil of clouds, casting a soft glow over the town that lay before Elias. He drew a deep breath, squaring his shoulders as he ventured out into the familiar streets, determined to bridge the gap carved by years and wars. It had been too long since he’d walked these paths, where each step echoed memories of a simpler time.

\textbf{chapter 2: Graves and Memories}

As he neared the heart of town, Elias couldn’t help but notice the gravestones lined neatly along the cemetery’s edge, their cold gray surfaces stark against the vibrant green of the grass. His heart clenched at the sight, emotions entwined like the gnarled roots beneath the soil. Some stones were new, bearing names freshly etched—bitter tokens of lives extinguished too soon. His gaze halted on a particular marker, and a sharp intake of breath caught in his throat.

Thomas. The name reverberated through his mind like a distant gunshot, unexpected and jarring. Clara’s brother, his childhood friend, his comrade in war. The flashbacks surged forth, a torrential flood of pain and chaos—the mud-soaked trenches, the cries of fallen men, Thomas’s wide eyes staring at a future forever lost. A paralyzing grief anchored him.

In that moment of overwhelming darkness, a stranger’s gentle touch found his arm. “Easy there, son,” they murmured softly, steadying him with an assuring presence. Elias nodded, grateful for the lifeline, even as his thoughts roiled like a storm.

Finally, Clara appeared near the cemetery, her footsteps measured, her expression unreadable. She approached Elias tentatively, a cautious warmth in her eyes that spoke volumes of shared history and unspoken sorrow. “Elias,” she began, her voice a fragile tether to the present. “I… I didn’t know you’d come back.”

He nodded stiffly, the words lodged deep in his chest, unable to meet her gaze. “I thought it was time.”

Their conversation was a delicate dance around the edges of grief. Each word was a petal fallen in a still pond, rippling outward but never reaching far enough to disturb the depth of their pain. Clara extended an invitation. “Would you like to come over for a bit? I could use the company. And I think you could too.”

The pretense of company thinly veiled a mutual need—the thirst for connection amidst desolation. Elias hesitated, guilt tightening his throat. How could he face her, knowing he had left Thomas behind on that forsaken battlefield?

Sensing his inner conflict, Clara added softly, “It’s not easy, being here without him. Thomas would’ve wanted us to be there for each other.”

Elias nodded, a silent acknowledgment of shared grief, and accepted her invitation, knowing it was also a step toward healing.

Over steaming cups of tea, Clara began to share stories of Thomas, each word carefully crafted to preserve his essence. Her laughter was brittle, yet sincere, an echo of a sibling bond that transcended the grave. Elias tried to contribute, but the letters twisted foreignly in his mouth, the weight of his own guilt threatening to crush him.

“Why didn’t you come back?” Clara’s question was seeded with plaintive hurt, the silence that followed more scorching than any answer Elias could muster. He struggled, barely glancing at her, knowing the potency of her gaze would unravel him completely.

“Clara…” he breathed, his voice strangled, a confession buried in the syllables.

She was not deterred by his reticence. Sitting beside him, her eyes mirrored a resolute resilience that inspired awe, even as it highlighted her masked distress. “You’re not alone in this,” Clara promised, reaching out, offering comfort wrapped in the warmth of her presence.

“I know,” he admitted, a small sigh escaping him as if acknowledging a cumbersome truth. “This town’s changed. Everything feels different.”

Clara nodded, understanding gilded with sadness. “But not everything is lost. Somewhere beneath the scars, there’s still life, waiting.” Her belief kindled a small light within Elias, one he hadn’t felt in a long time.

They decided on a walk, as Clara suggested, an old pastime now renewed with poignant intent. Elias’s reluctance to plunge back into the memory-laden streets wavered with her encouragement. He found himself accepting her proposal, their strides eventually syncing to a rhythm, solitary yet unified in silence.

Elias observed Clara, noticing her guarded demeanor, and in the ensuing quiet, he reflected on her resilience, her steadfast refusal to be consumed by sorrow. It was inspiring, a reminder that survival could be more than mere existence.

As they walked, Clara brushed against Elias’s hand, her own trembling slightly, betraying the fear she hadn’t voiced. “I can’t bear to lose anyone else, Elias,” she revealed, the fear of isolation looming ever large. Her vulnerability prompted a surge of protectiveness in him.

Pausing at an ancient oak, its branches casting dappled shadows on the ground, Elias found his voice. “Clara, I… I can’t promise much. But I’ll try to be here, try not to run.”

A quiet smile unfolded on her lips, appreciation mingled with relief. “Then that’s enough.” There was an understanding in her words, straightforward yet profound, a pact of endurance.

As they prepared to part, Clara squeezed his hand once more. “Promise me, you’ll come by more often?” Her insistence was gentle yet firm, her determination to keep him anchored starkly apparent.

Elias met her gaze at last, the weight of shared tribulations fostering a new connectivity. “Yes, I will. I promise.”

They lingered a moment longer, enveloped in the unspoken promise of better days. And as the sun dipped below the horizon, Elias felt the stirrings of hope—a fragile, tentative thing, yet it was growth nonetheless.

\textbf{chapter 3: A Child's Light}

The late afternoon sun wove golden threads through the patches of trees lining Clara’s street as Elias approached her modest home once more. The house itself stood humble, adorned with a faded blue door and a small garden spilling fragrant petals onto the cobblestones. Elias paused at the gate, his fingers ghosting over the iron latch. His chest tightened, but a deep breath steadied him—a momentary anchor against the churn of memories and self-doubt.

Clara greeted him at the door, her smile soft but welcoming, betraying none of the tension she carried. "Come in," she beckoned gently, stepping aside to let him into the cozy warmth of her world. The living room was much the same as he remembered—cluttered yet homey, filled with books, knickknacks, and the faint smell of vanilla. But today, the space seemed brighter, animated by the cheerful hum of a child’s voice.

“Elias, this is Sophie,” Clara introduced, gesturing to the young girl perched cross-legged on the living room rug, a pair of large, curious eyes lifting to meet his. Sophie, barely six years old, had an untamed halo of curls and a demeanor as radiant as the sunbeam that danced across her shoulders.

“Hello!” Sophie chirped, bounding up to him with the unreserved energy of childhood. “Are you my Aunt Clara’s friend? Do you want to play with me?”

Elias blinked, startled by the tiny whirlwind now tugging at his sleeve. He glanced at Clara, who nodded encouragingly, clearly amused by her niece’s enthusiasm. "Sophie, let’s give Elias a moment to settle in," Clara admonished lightly, though the twinkle in her eye betrayed no real urgency to deter Sophie’s exuberance.

The girl pouted but relented, retreating to her game of make-believe yet keeping Elias under watchful scrutiny as he took a hesitant seat near the fireplace. Moments later, though, her persistence proved boundless. She returned, clutching a small tiara made of plastic. "You can’t say no to a princess, you know," Sophie declared solemnly, thrusting the tiara toward him.

Elias faltered, caught between awkwardness and the faintest stirrings of amusement. "I suppose not," he conceded, lowering himself to the rug. Sophie’s beaming smile made it worth the tentative step out of his comfort zone. She explained her make-believe world with the earnestness of a royal decree, pulling him into her realm of castles and queens and talking foxes. Though the weight in his chest lingered like an unwelcome guest, it receded ever so slightly—just enough for the edges of a small, unfamiliar smile to creep in. Clara watched from the doorway, her arms crossed and an indescribable mixture of hope and relief softening her features.

At one point, Sophie clapped her hands suddenly and announced, "It’s your turn to share a story!" Her wide eyes bore an unspoken challenge, one Elias found surprisingly difficult to refuse.

“I don’t know any stories," he muttered feebly, causing Sophie to gasp as though he’d committed a terrible crime.

“Everyone knows stories! Even Uncle Thomas told me stories before he… before he went away," Sophie said, her voice dipping briefly into a quiet valley.

The room fell still, Clara’s gaze pinning Elias, her expression layered with complexity. He hesitated, the name surfacing like a jagged shard lodged in his throat. But Sophie’s eager anticipation pushed him forward. “All right," he relented, his voice low but steady. "Once, long ago, there was a soldier…" As he spun a simple tale for Sophie, lacing it carefully with imaginary adventures rather than harsh truths, her laughter began to mingle with the words, filling the room like spring music. And though it shouldn’t have, her joy coaxed a laugh from him—a rare, unguarded sound Clara hadn’t witnessed in years.

The moment hung in the air like a fragile orb, beautiful in its impermanence. Clara’s eyes shimmered faintly as she folded her hands together. "He would’ve liked seeing this," she murmured, mostly to herself. And then, louder: "Elias, would you stay for tea?" Her tone carried a quiet plea, and Elias knew it wasn’t just tea she was asking him to linger for. He nodded, and her smile deepened.

Throughout the afternoon, Sophie’s unrelenting curiosity turned to questions about Elias’s life, about soldiers, about his brother. Her innocence was jarring, yet strangely grounding. "Do you miss Thomas?" she asked suddenly, peeking up from her drawing of a princess riding a dragon.

Elias froze, his eyes drifting to the medallion tucked beneath his shirt. But instead of retreating into himself, as he so often did, he knelt beside Sophie. "I do," he answered honestly, his hand resting lightly on her drawing. "Every day."

She nodded solemnly, as though she somehow understood the weight of his words. Then she held the paper out to him. "You can help me color. Red for the dragon’s tail. It’s the strongest color."

Something shifted in Elias as he eased into the calmness of her perspective. Red wasn’t just for blood and loss. Red could be strength, bravery—even hope.

Clara moved quietly through the house, setting the table and watching as the layers of Elias’s guarded demeanor began to peel back ever so slightly. She paused by the window, her expression unreadable as she turned something over carefully in her mind. Finally, carrying a plate to the table, she spoke. "Would you like to share something about Thomas?" Her voice was a delicate waver, attempting to pull Elias into deeper waters without drowning him.

Elias’s eyes flicked toward her, then to Sophie, who was busily applying glitter to her drawing. For a moment he felt fear—raw, gnawing. But Clara’s steady gaze rooted him. "Thomas," he murmured, almost tasting the name. "He… he gave me this," he added, reaching into his pocket. He revealed the medallion and handed it to Sophie, who inspected it wordlessly before handing it back with an approving smile.

Later, Sophie fetched a small gift from her room and pressed it into Elias’s hand—a bracelet of woven threads in mismatched colors. "It’s for you! So you don’t forget."

Elias chuckled softly. "Forget what?"

She frowned at him playfully. "That you’re supposed to smile sometimes!"

The sun dipped lower as dinner drew near. Elias helped Clara set the table at her suggestion, his movements clumsy but earnest. The sound of Sophie’s laughter mingled with the clatter of dishes, and for a brief moment, the house felt alive in a way Elias hadn’t known he’d been missing.

As the evening wound down, Sophie insisted he choose a favorite storybook for her bedtime reading, though her excitement left her asleep halfway through. Clara returned to the quiet living room where Elias sat, the bracelet still in his hand. "She’s taken quite a liking to you," she said with a smile.

Elias exhaled, his own small smile playing at the corners of his lips. "She’s hard to say no to."

As he prepared to leave, Clara walked him to the door. “Will you come back for dinner? Next week?" she asked hesitantly, gauging his reaction. For a heartbeat, Elias hesitated, but then he nodded. 

“I’ll come back," he promised softly, the words feeling less like a burden and more like a bridge.

Clara’s smile broadened, radiating a gratitude that illuminated the darkness between them as Elias stepped back into the night, carrying the faint glow of the tiny beacon named Sophie.

The morning air was filled with the smell of sizzling bacon and freshly brewed coffee as Elias made his way to the kitchen, the familiarity of home wrapped around him like a well-worn coat. Yet, as he entered the room, an undercurrent of tension flickered just beneath the surface, unseen but palpable. His father sat at the table, a sturdy figure hardened by years of labor and expectations, his face shadowed by the newspaper he seemed more interested in than the breakfast his wife faithfully laid before him.

"Elias," his father began, not bothering to look up from the news, "what are your plans now that you're back home? Have you considered a position at the factory? It's a good, honest living, just like it has been for us Merricks."

Elias hesitated, the fork in his hand suddenly feeling like a foreign object. His father’s voice carried both the weight of tradition and the unspoken demand for his obedience. "I... I haven't decided yet," Elias mumbled, focusing his gaze on the plate before him. Yet, his evasiveness did little to stop the burgeoning storm.

"Haven't decided?" His father's voice rose slightly. "Elias, this town is trying to recover, and you're needed here. You've been hiding behind your war stories long enough. What are you waiting for?"

The words struck like blows, each syllable etching deep into the fragile surface of Elias’s composure. "You think I’m hiding?" Elias shot back, his voice breaking the calm he had desperately tried to maintain. "You think I’m using my trauma as an escape? It isn't that simple, father. You don't understand what I've seen, what I've been through."

His father slammed the newspaper onto the table, eyes blazing with unresolved frustration and something deeper, something more akin to fear. "Maybe I don’t understand, but that doesn’t change the fact that we need every hand we can get here, and you—"

“Enough!” Elias's voice cracked like a whip, sending a silence rippling through the room. His mother, always the calming presence, stepped between them, her hand a gentle plea resting upon his father’s shoulder.

“Let's not do this now,” she whispered softly, her eyes searching Elias’s face for something she couldn’t quite find. "We can talk about it later, after we've all calmed down."

The damage was done, though. The air felt thick with the unsaid and the unhealed. Elias’s heart hammered in his chest as he stepped back, shaking off the layers of disappointment and rage that clung to him. Without another word, he turned and left, the door closing behind him with a final, hollow thud.

The path leading to the riverbank unraveled beneath his feet, memories clinging like shadows to each familiar bend. Thoughts of Thomas and their time here drifted through his mind, stinging with the bittersweet reminder of simpler days. Once upon a time, this had been their sanctuary, a space untainted by the complexities of family and expectation.

As Elias sank onto the damp earth by the river, he caught sight of something partially buried in the underbrush—a worn journal, its cover stained and frayed. Thomas’s old journal. With a sense of reverence and a heavy heart, Elias picked it up, thumbing through pages filled with sketches, notes, and snippets of dreams that belonged to a world before everything changed.

Thomas’s voice seemed to rise from the pages, whispering hopes and fears that felt painfully close to Elias’s own tangled thoughts. He read through it, tears slipping unbidden from his eyes as the memory of his brother’s laughter, his unwavering belief in Elias, flooded back. For a moment, Elias wept openly, mourning the loss of two brothers—the one buried beneath foreign soil and the one he once had been. 

Eventually, Elias’s gaze drifted back up to his surroundings, the sun casting shimmering reflections on the river's surface. Nearby, he noticed old carvings etched into the trees—the handiwork of a much younger father and son. He traced the familiar patterns, feeling the aching distance between the father he remembered and the man he now had to face.

For the first time in weeks, amidst the rustle of leaves and the gentle babbling of the stream, Elias found a sliver of peace. He picked up a piece of driftwood and began to carve. The act was meditative, his hands moving instinctively as he worked the wood, transforming it into a vessel for words he couldn't quite say aloud.

The little driftwood transformed under his hands, a silent letter to a father who spoke love in deeds rather than words. Elias etched into it first his father's old symbol, thoughtlessly carved into the bark many seasons ago, now recreated with the deliberate care of a son seeking reconciliation. 

When he finished, Elias held the piece up to the fading light, watching the play of shadows across its surface. The message was simple but heartfelt: an apology, a symbol of love, and a commitment to try and bridge the chasm forged between them.

As the sun sank further behind the horizon, Elias resolved to return, prepared for the difficult conversations that awaited. He placed the driftwood down on his father’s workbench, allowing the gesture to speak where words might falter.

The return home was slow, his mother waiting by the door, her silhouette softened by the night’s gentle embrace. "He doesn’t mean to hurt you," she said softly, drawing him into the warmth of the kitchen. "His words come from a place of love and fear for you, Elias. Just as they always have."

Elias nodded, the remnants of his earlier anger fading into a quiet resolve. "Maybe," he replied, though uncertainty still twisted through his heart. But beneath the ambivalence lay a renewed sense of purpose—one that promised to guide him not away from, but towards the family he had stood apart from for too long.

\textbf{chapter 4: Voices in the Hall}

The sun dipped low on the horizon, draping the cobblestone streets of the town in hues of burnt orange and soft violet. Elias shifted uneasily on the bench at the edge of the square, his hands gripping his knees as if bracing himself for the weight of moving forward. Across from him, the town hall loomed—an unassuming building with whitewashed walls and windows that spilled golden light onto the darkening street. Through the open doors, the faint murmur of voices reached him, a low hum of conversation mingling with the occasional burst of laughter or a lingering note of sorrow. 

Clara stood beside him, her presence both grounding and insistent. She had the kind of patience that felt neither dismissive nor overbearing, but the determination in her eyes left little room for retreat. "Elias," she began softly, lowering herself to his level, "this isn’t about having all the answers or knowing what to say. It’s about being here, in this moment. You. Deserve. To. Be. Here."

He exhaled slowly, the weight of her words pressing on him in ways both comforting and heavy. "I don’t know if I can," he muttered, staring at his boots. Thoughts of his father’s voice, sharp and full of expectation, clashed with the quieter echoes of Thomas’s laughter—two specters that fought for space in his mind. 

"You can," Clara said, her voice steely. "And you will. You’ve survived far worse. All I’m asking is that you walk through those doors. Just to listen. That’s all."

Elias hesitated, her words lingering in the air like a melody he couldn’t quite grasp. But something about the way she said it, something about the unwavering belief in her tone, stirred a flicker of resolve in him. At last, he nodded, though his throat felt tight with apprehension. "Alright, Clara. I’ll go."

The warmth of her smile eased some of the tension in his chest, and with that, she led the way.

Inside, the hall was alive with quiet energy. Rows of folding chairs faced a modest wooden podium. Faces of all ages filled the room—some familiar from years past, others strangers tethered together by invisible threads of loss, love, and resilience. The air smelled faintly of old varnish, with the occasional tang of coffee cups grasped tightly in nervous hands. 

They slipped into seats at the back, Clara offering a small nod to the people they passed. Elias sat stiffly, his shoulders square and his breaths shallow, scanning the room like a soldier surveying a battlefield. Voices rose and fell as those gathered greeted each other, shared soft words, or simply offered silent nods of acknowledgment. 

The first story began with a widow—an older woman in a muted blue shawl, her hands clasped tightly around a handkerchief as if to anchor herself in the storm of her own words. Elias couldn’t stop himself from listening. Her voice was unsteady but full, each word trembling under the weight of memory. She spoke of her husband, a kind man who had carried compassion as naturally as others carried burdens. She told the hall about his quiet acts of service—a loaf of bread left on a struggling neighbor’s doorstep, a night spent fixing a leaking roof in a storm. Her tears came freely, but her voice never broke, her grief braided with an unmistakable pride. "He wasn’t a perfect man," she finished, "but he was good. And sometimes, goodness is harder to hold onto than perfection." 

Elias felt a knot loosen in his chest. He didn’t know the woman’s husband, and yet, he felt as though he did. Her words were a reminder that even in the face of overwhelming loss, lives could leave echoes that warmed the spaces they once filled.

Clara leaned closer to him. Her elbow brushed his, a subtle gesture of encouragement. "People are brave in their own ways," she whispered. "Even showing up is brave."

But Elias said nothing. His gaze shifted across the room, where another voice had begun—a soldier this time, young and wiry, his uniform clean but ill-fitting, as though he hadn’t yet grown into its weight. The soldier’s story was raw, his words laden with the edge of recent wounds. He spoke of an ambush, of running through a forest thick with enemy fire, of a comrade who had thrown himself on a landmine so the others could escape. Silence fell heavily in the room as he described the agonizing crawl to safety, his throat dry from thirst and his body screaming in protest with every inch gained.

Elias found himself leaning forward, his fists tightening in his lap. He felt the soldier’s fear in his own chest, the echoes of his memories bouncing painfully against his ribs. Yet, when the soldier paused to collect himself, the room did not judge his hesitation. Instead, there was an outpouring of warmth—murmured words of gratitude, a light touch on the shoulder, nods of understanding. Elias couldn’t look away. He noticed how the young man sat straighter after that, his trembling hands steadied by the quiet strength of those around him.

The stories wove threads through Elias. He observed families in mourning—their grief worn openly, their smiles brittle but alive. He saw a young woman gripping a photograph of her brother so tightly that her knuckles turned white. He caught the nervous tremor of a speaker’s voice as they stammered through tearful lines, and the resolute applause that followed, as if to remind them they were seen and heard. The vulnerability laid bare before him wasn’t something he had anticipated. It was raw but beautiful, each shared memory a piece of a greater mosaic, every tear a testament to bonds that outlived pain.

As the night carried on, Elias wrestled inwardly. Memories of his own clung to him like shadows—Thomas stealing bread to share with starving children, the laughter of privates beneath a tattered tent, the cold weight of dog tags pressed into his palm. He tried to picture himself standing at that podium but couldn’t imagine the words spilling forth. Even the thought left his throat dry, his chest tight. What could he possibly say? How could he begin when everything felt so fractured?

Clara nudged him again, sensing his inner turmoil. "You don’t have to speak tonight. Just think about it," she said gently. "It’s a process, Elias. It’s not about perfection. It’s about connection."

He nodded slowly, her words coiling around his thoughts like vines. As the evening concluded, applause followed the final story—a young boy speaking haltingly about his father’s letters from war, full of wisdom and love sent from a battlefield he never returned from. Elias felt tears brimming in his eyes, though he left them unshed.

Walking out into the cool night air, he inhaled deeply, the silence between him and Clara filled with unspoken gratitude. His steps felt lighter. She squeezed his hand as they reached the bench, her face glowing softly in the moonlight. "You were brave tonight, just by being there," she said. "And that’s enough."

Elias turned to her and managed a small smile. "Thank you," he said, his voice low but steady. For the first time in a long time, he let himself imagine a future where his silence didn’t hold him hostage. The seed Clara had planted within him stirred faintly, a quiet promise waiting to bloom.

\textbf{chapter 5: Teaching Tomorrow}

The morning air was crisp as Elias made his way down the narrow path lined with dewy grass, his thoughts clouded by memories and unspoken fears. The schoolhouse stood modestly at the end of the lane, its brick walls washed with the early light of dawn, casting long shadows that seemed to reach out to him. It was here, in this quaint place of learning, that Ms. Hargrove had made her unexpected request.

She had found him by the river a few days prior, where he often retreated to battle the noise in his mind with the gentle rush of water. Her approach had been cautious, her voice placid as she posed her question. “Elias, I wanted to ask if you would consider speaking to the students about the war.”

His reaction had been immediate and visceral. “No,” he had snapped, the word sharp and cutting. He didn’t have the strength to drag the ghosts of his past into the light, to let them dance across the minds of children.

But Ms. Hargrove had not flinched, her gaze steady and understanding. “Think about it,” she urged softly before leaving him to his thoughts.

The days that followed were a blur of doubt and persuasion. Clara, ever his steadfast companion, spoke to him with a gentle resolve. “Elias, your story matters. The children need to hear about both the heroism and the sacrifice, so it doesn’t repeat itself.” Her conviction simmered in her words, melting some of the ice around his heart.

Elias struggled with his fear, the idea a haunting specter, mocking his silence. He wrestled with himself through sleepless nights, pacing the creaky floorboards of his small home as his memories taunted him with whispers of broken promises and lost futures. Yet, slowly, Clara’s reasoning pierced through his defenses, and with a tentative nod, he found himself agreeing to visit the school.

Preparing for the talk was like wading through thick fog, each step heavy with uncertainty. He rehearsed his speech alone, the words feeling foreign in his mouth, each syllable a reminder of what he had witnessed, what he had lost. Despite the trembling fear that gripped him, he persevered, knowing Clara’s belief in him was a lifeline he could not ignore.

The morning of the visit arrived too soon. As he stood in front of the school, the anxiety washed over him in waves. But inside, Ms. Hargrove was there, her presence a calming anchor amidst the storm. She introduced him to the classroom, her manner polite yet encouraging.

Elias faced the students, their wide eyes and expectant faces staring back at him. His voice stammered as he began, each word a hill to climb. “I’m here to share a part of my life… a part of many lives,” he managed, feeling raw and exposed.

He shared his war experiences, his voice faltering like a fragile bridge spanning the gap between past and present. The students listened, their attentiveness a balm to his worn spirit. As he recounted the tale of Thomas’s bravery, the classroom blurred, a constellation of emotion and memory.

The room was silent, save for the occasional shuffle of feet or rustle of paper. Encouraged by the quiet empathy that surrounded him, Elias found himself breaking down emotionally, tears cutting paths down his cheeks.

Surprisingly, the students didn’t recoil but instead asked questions—some thoughtful, others tinged with innocent curiosity—drawing him further into the conversation rather than pushing him away. Their earnestness stirred something in him, allowing him to acknowledge the support surrounding him.

In the uneasy quiet that followed, Elias reflected on his personal loss, how it had shaped the man he was today. He realized, as he stood before the children and the specter of his past, that his grief was not a solitary burden but interconnected with the community’s shared experiences. 

The students’ empathy glimmered in their eyes, a halo of unexpected courage lighting within Elias. He found himself standing taller, his voice gaining strength with each revelation.

A moment of clarity washed over him like a fresh tide. The vulnerability he feared had instead rendered him free, reminding him that his scars spoke of endurance, of lessons that held the potential to shape a different future.

As he concluded, the classroom was filled with a gentle silence, a shared understanding that transcended words. Ms. Hargrove led the applause, her face glowing with pride.

Elias embraced this newfound vulnerability, feeling the warmth of the students’ appreciation and the gentle surge of bravery in his chest for the first time since the war. 

He left the school with a lighter step, the sky opening up in endless blue above him. The weight he had carried was still there, but lighter now, as if others shared in its burden. For the first time, Elias allowed himself to hope for healing, to believe that one day, he might find peace.

\textbf{Chapter 6: Rebuilding Together}

Elias had never imagined himself as a community activist, yet here he was, standing in the brisk morning air, ready to join hands with the people of his town. It was an almost surreal shift, inspired by the realization that change started with small yet meaningful steps. The community group had their headquarters in an old barn, the roof slightly sagging but otherwise sturdy. As he entered, a warmth that rivaled the cold sunshine outside greeted him.

The room buzzed with energy as volunteers gathered, their chatter creating a symphony of optimism. With his heart beating with renewed purpose, Elias approached the group leader, a burly man with a kind face named Marcus. ‘Welcome, Elias,’ Marcus said, clapping a big, reassuring hand on his back. ‘Glad to have you with us.’

Elias soon found himself paired with a few other volunteers, Clara among them. ‘Looks like we’re teammates again,’ she teased, her smile wide and infectious. Together, they worked on their first project: repairing the run-down community center. The building was an eyesore, its once vibrant walls dull and chipped. Yet, in its bones, Elias sensed the stories it held.

As they began stripping away the old paint, Clara and Elias worked side by side, their movements synchronized in harmonious rhythm. Hours passed, sweat running rivulets down their faces, but they pressed on, driven by the vision of what the community center could become.

Midweek, Clara paused in her labor, turning to Elias with a gentle urgency in her eyes. ‘I think you should talk at the town meeting,’ she suggested. Her encouragement was unwavering, as ever. ‘You have a voice, Elias, and people are listening.’ The prospect terrified him, but he nodded, realizing her belief in him was something to trust.

By the week’s end, Elias attended the meeting, his voice understated yet compelling as he shared his thoughts. The room listened, the air thickening with hope that bubbled beneath the surface.

Days later, as Elias was hammering the last nail at the work site, his father appeared. Their eyes met, and the silence that unfurled between them spoke volumes more than words could. ‘Need some help?’ his father asked, uncharacteristically shy. The gesture was promising, a tenuous olive branch.

Later, as the evening sun bleached the sky, Clara shared a story from her childhood, their laughter mingling with the rustling leaves overhead. ‘I used to dream about painting this town in colors of hope,’ she confessed with a soft chuckle. Inspired, they coordinated efforts to bring her dream to life, discovering a forgotten mural in the community center. Its faded colors revealed glimpses of history—a tapestry of the town’s spirit.

Elias and Clara organized a community meal, their excitement tangible as they crafted dishes together, each ingredient infused with shared camaraderie. That night, as fireflies waltzed in the garden outside, Clara and Elias found themselves painting the recovered mural, restoring its glory stroke by stroke. It was a labor of love.

Rebuilding efforts gathered momentum, and soon, Elias and Clara were knee-deep in salvaging materials, driven by the clear vision of a shelter they planned to build for a displaced family. Alongside the teenagers, they became mentors, guiding them through projects that impressed upon them the spirit of community.

Their days were filled with the chaos of planning events and meetings, Elias frequently finding respite in quiet moments of reflection with Clara. One evening, as they gazed at the renovated playground, their expressions mirrored the quiet satisfaction of witnessing tangible progress.

Comforted by the bond effortlessly forged amid shared toil, Elias confided in Clara about his father. ‘I think he wants to change, like I am,’ he admitted, his voice laced with thinly veiled hope. Clara’s supportive presence was a constant, her silent understanding grounding him.

As the rebuilt swings in the playground screeched into life, they bore testament to the labor and dreams invested by both Elias and Clara. Together, they even began discussing future projects, eager to continue their mission.

The work never ceased, but exhaustion was a shadow that brought them closer, their resolve strengthened by the shared burden. As they strolled home one evening, the town lit by the blush of twilight, a silent understanding flowed between them, their unspoken emotions simmering beneath their tranquil expressions.

As they reached the corner where their paths diverged, Elias paused, his gaze meeting Clara’s under the streetlights casting dappled shadows. Their roles in rebuilding were more than tasks; they were a testament, a bridge to something profound and enduring. For Elias, amidst community connections and newfound purpose, the journey to heal was just beginning, and he was grateful not to walk it alone. Together, they faced the burgeoning promise of tomorrow.

\textbf{Chapter 7: Seeds of Hope}

Elias stood at the edge of the town square, where the morning sun gently draped its light over cobblestones and quaint shops, breathing life into the awakening town. Today felt different. It was not just another day, but a day bearing the weight of promise. As he stared at the modest setup—a small stage and rows of chairs facing it—he took a deep breath. He had organized his first community event, a gathering rooted in hope and growth: a sapling planting ceremony.

The early spring breeze carried the rich, earthy scent of freshly dug soil and the muted chatter of townsfolk arriving, curious and eager. This moment was months in the making, born from the embers of conversations and dreams shared over paintbrushes and community dinners. Yet, beneath the organizational tasks lay Elias’s personal journey—a quest for healing and connection, now materializing as this budding endeavor.

As more people filtered in, Elias, adorned in a simple yet neat attire, stepped onto the stage. When the chatter quieted, he cleared his throat, scanning the faces before him. Among them, veterans, carrying stories etched in their lined faces; widows, holding memories in their clasped hands; children, with eyes wide open to promises yet unfulfilled, and there in the crowd, his parents, standing together, their presence a balm of acknowledgment and acceptance.

With a steadying exhale, Elias began his speech. ‘We gather today not just to plant a sapling, but to sow seeds of resilience,’ he spoke, his voice warmed by the sincerity of his convictions. ‘This sapling stands as a symbol of new beginnings, a testament to our strength in the face of adversity. Though our scars are deep, they mark us as survivors.’ His gaze drifted across the crowd until it landed on Clara, her presence infusing him with strength.

As the speech resonated among the crowd, Clara stepped forward, her eyes meeting his with a strength that spoke volumes. The townsfolk joined her, each taking solace in the shared mission of fostering new life. Sophie, the ever-kind presence in their efforts, offered a bouquet of wildflowers, sweet and vibrant, to adorn the budding sapling.

A child inched closer, tugging on Elias's jacket with curiosity. ‘Why do we plant trees?’ he asked, eyes reflecting the innocence of simple wonder. Elias smiled, crouching to meet the boy’s gaze. ‘Because trees help us remember that no matter how hard the past has been, we can always grow again,’ Elias acknowledged, offering truth wrapped in hope.

An elderly veteran, standing tall with dignity drawn from decades of resilience, shared a fragment of his own history—the war, the return, and the finding of peace in everyday routines. His words were threads weaving a rich tapestry of shared experiences, binding the community in mutual understanding.

The ceremony reached its gentle climax as Clara knelt beside the awaiting hole for the sapling. Her hands tenderly cradled the young tree, symbolizing the nurturing act they were all partaking in. Elias noticed the gesture and invited Clara to stand with him, acknowledging her as not just an ally in this project, but a partner in its emotional and spiritual journey.

Elias guided the young child towards the sapling, helping him to add his handful of soil, a small yet significant act. The townsfolk erupted in applause, a sound as invigorating as spring rain, appreciating the efforts of this young man who had steadily grown into a figure of inspiration among them.

Gesturing towards the sapling, Elias spoke with reverence. ‘This is not just a tree—but a beacon. May it inspire us to nurture our roots and reach for the sky.' Sophie, taking Elias's cue, led the children in a line, each awaiting their turn to contribute.

As the day waned and sunlight danced through budding leaves, Elias took a moment to reflect on the improbable journey that had led to this luminous present. There was a palpable shift, a warmth that extended beyond the sun’s reach, wrapped in the company of those he cared for.

Clara's hands met the sapling with a wellspring of kindness, the silent act as significant as any grand gesture. Elias turned towards the townsfolk, offering heartfelt thanks for their solidarity, their shared journey in this healing process. Though words of encouragement, he invited everyone to reflect, to see in this humble tree the resilience within themselves.

With deliberation, Elias placed soil around the base of the sapling, its future now intertwining with theirs. He caught Clara’s eye, a gentle exchange passing between them—an unspoken acknowledgment of challenges faced and a shared, hopeful horizon.

‘Let’s nurture this together,’ Elias encouraged, addressing the crowd, envisioning a collective effort to protect and watch the sapling grow, just as they would with each other.

As the townsfolk lingered, sharing stories old and new, Elias found solace on a nearby bench, Clara and Sophie settling beside him. The three sat in comfortable silence, the sapling standing proud under the dimming sky, a still tender but promising symbol of growth. And though Elias knew the road ahead was filled with trials, surrounded by friends and memories rooted in healing, he embraced the bittersweet beauty of moving forward, one tender step at a time.

}
\end{spacing}

\end{tcolorbox}

\end{document}